\documentclass[5p,times]{elsarticle}

\usepackage{lineno,hyperref}
\modulolinenumbers[5]

\usepackage{wrapfig}

\usepackage{multirow} 
\usepackage{booktabs}
\usepackage{amssymb} 
\usepackage[ruled,linesnumbered]{algorithm2e}
\usepackage{graphicx}
\usepackage{verbatim} 
\usepackage{makecell}
\usepackage{amsmath}     
\usepackage{bm}
\usepackage{booktabs,xcolor,colortbl}
\usepackage{xparse}
\usepackage{float}
\usepackage{subfig}
\usepackage{parskip}

\usepackage{amssymb}
\usepackage{amsthm}
\usepackage{txfonts}

\usepackage{xcolor}

\journal{Journal of Neurocomputing}

\begin{document}

\begin{frontmatter}

\title{HIRE: Distilling High-order Relational Knowledge From Heterogeneous Graph Neural Networks}

\author{Jing Liu 
\corref{co-first authors} \fnref{mymainaddress,mysecondaryaddress}}
\ead{liujing18z@ict.ac.cn}

\author{Tongya Zheng
\corref{co-first authors}
\fnref{cs}}

\author{Qinfen Hao
\corref{correspond}
\fnref{mymainaddress}}
\ead{haoqinfen@ict.ac.cn}

\cortext[co-first authors]{These two authors contributed equally to this work.}
\cortext[correspond]{Corresponding author}

\address[mymainaddress]{SKLP, Institute of Computing Technology, Chinese Academy of Sciences, Beijing, China}
\address[mysecondaryaddress]{University of Chinese Academy of Sciences, Beijing, China}
\address[cs]{College of Computer Science and Technology, Zhejiang University, Hangzhou, China}

\begin{abstract}
   Researchers have recently proposed plenty of heterogeneous graph neural networks (HGNNs) due to the ubiquity of heterogeneous graphs in both academic and industrial areas. Instead of pursuing a more powerful HGNN model, in this paper, we are interested in devising a versatile plug-and-play module, which accounts for distilling relational knowledge from pre-trained HGNNs.
    To the best of our knowledge, we are the first to propose a \textbf{HI}gh-order \textbf{RE}lational (\textbf{HIRE}) knowledge distillation framework on heterogeneous graphs, which can significantly boost the prediction performance regardless of model architectures of HGNNs. Concretely, our HIRE framework initially performs first-order node-level knowledge distillation, which encodes the semantics of the teacher HGNN with its prediction logits. Meanwhile, the second-order relation-level knowledge distillation imitates the relational correlation between node embeddings of different types generated by the teacher HGNN.
	Extensive experiments on various popular HGNNs models and three real-world heterogeneous graphs demonstrate that our method obtains consistent and considerable performance enhancement, proving its effectiveness and generalization ability.
\end{abstract}


\end{frontmatter}

\section{Introduction}\label{sec:introduction}
\begin{figure}[h]
	\centering
	\includegraphics[width=0.5\textwidth]{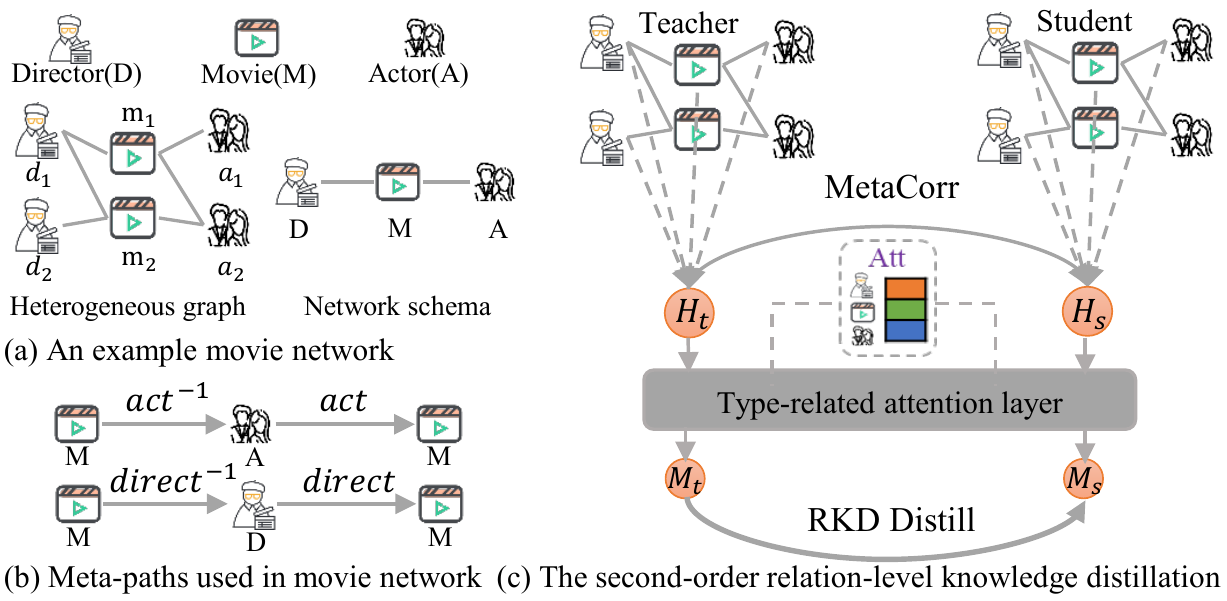}
	\caption{Figure(a) is a toy example of a heterogeneous graph for the IMDB dataset. It consists of three node types (i.e., movie, actor, director) and two edge types (i.e., $direct$, $act$), involving two meta-paths (MAM and MDM) shown in Figure(b). Here Movie (M) is the target node type related to the downstream task of predicting the movie genre. Figure(c) is an overview of the crucial part of our proposed HIRE, where MetaCorr establishes the correlation between different types of nodes, Att is to highlight the importances of different nodes, and RKD is to distill the second-order relation-level knowledge between node embeddings of different types generated by the teacher HGNN. 
	}
	\label{fig:example}
\end{figure}

    Heterogeneous graphs, which contain various types of nodes and relations, are ubiquitous in the real world, such as academic networks, movie networks, and business networks~\cite{wang2022survey,shi2016survey}. For example, IMDB in  Figure~\ref{fig:example}(a) contains three types of nodes: actors, movies, and directors, as well as different types of relations between them (see Figure~\ref{fig:example}(b)). Learning node representation in heterogeneous graphs is important for numerous graph analysis applications, such as recommendation~\cite{shi2018heterogeneous,zhao2017meta}, text classification~\cite{linmei2019heterogeneous,kilimci2018deep}, financial risk management~\cite{liu2018heterogeneous,gutierrez2020multi}, Biomedicine~\cite{bongini2021molecular,yu2021resgnet}, and traffic forecasting~\cite{lu2020lstm,zhou2021ast}.

	Heterogeneous graph modeling, compared with homogeneous graphs, has the advantage of integrating more information. Nonetheless, it incurs a critical challenge of embedding the rich structure and semantic knowledge of heterogeneous graphs into low-dimensional node representations. 
	Recently, many GNNs-based methods have been proposed to tackle the heterogeneity of nodes and edges on heterogeneous graphs, and can be mainly divided into metapath-based and edge-based methods. 
	To capture the edge heterogeneity, the edge-based methods~\cite{schlichtkrull2018modeling,hu2020heterogeneous,yu2020hybrid} directly leverage the relation-specific matrix to deal with various types of edges in different metric spaces. However, the edge-based methods can only capture local structural information of heterogeneous graphs. 
	As an effective tool for semantic mining, meta-path~\cite{Sun2011PathSimMP}
	is proposed to capture more complex and richer high-order semantic information in heterogeneous graphs between nodes of the same type.

	Although the existing HGNNs have achieved promising performance, their representation capabilities are limited by: 
	(1)	Imprecise data labeling work. Generally, HGNNs are trained for semi-supervised learning, relying on extensive and high-quality labels of heterogeneous graphs. However, the ground-truth labels of target nodes are often selected from a list of true labels, making data labeling imprecise. 
	(2) Semantic relation modeling between different types of nodes. 
	Even though meta-path is designed to model high-order semantic, meta-paths selection for different fields is still challenging because it requires sufficient domain knowledge.

	To address the imprecise label problem, 
	knowledge distillation (KD) ~\cite{hinton2015distilling} enforces the student model to learn soft labels of the pre-trained teacher model instead of the one-hot label of nodes.
	It could establish the inter-class and intra-class correlation between the classes beyond the traditional one-hot labels.
	Inspired by KD, we introduce a \textbf{N}ode-level \textbf{K}nowledge \textbf{D}istillation (\textbf{NKD}) method to transfer the soft labels of the target node (e.g., Movie in movie networks) to students. As demonstrated in Figure~\ref{fig:framework}, NKD can provide general supervision information for downstream tasks such as node classification.

	To tackle the semantic modeling problem, it is critical to mine the correlations between heterogeneous nodes and edges, supervising the student model with the high-order semantics.
	To bridge this gap,  we design the \textbf{R}elation-level \textbf{K}nowledge \textbf{D}istillation (\textbf{RKD}), using semantic information inherent in HGNNs as another knowledge source. Primarily, we devise a correlation matrix named MetaCorr to encode relation-level knowledge between different types of nodes from pre-trained HGNNs, which accounts for distilling relational knowledge, as illustrated in Figure~\ref{fig:example}(c).

	In summarization, we propose a novel \textbf{HI}gh-order \textbf{RE}lational (\textbf{HIRE}) knowledge distillation method to apply KD to heterogeneous graphs, which distills both node-level and relation-level knowledge, as drawn in Figure~\ref{fig:framework}. Based on the prediction logits obtained by pre-trained teacher HGNN, we apply NKD to transfer soft labels to students. Given the relational correlation between node embeddings of different types generated by the teacher HGNN, we utilize RKD to transfer the second-order relation-level knowledge to students. By integrating NKD and RKD, HIRE can transfer the first-order node-level soft labels and the second-order relation-level correlation scores between different nodes, resulting in a practical and universal training approach that benefits various HGNNs.

    To validate the effectiveness of HIRE, we perform extensive experiments on ACM, IMDB, and DBLP datasets of node classification and node clustering tasks. Experimental results on RGCN~\cite{schlichtkrull2018modeling}, HAN~\cite{wang2019heterogeneous}, HGT~\cite{hu2020heterogeneous} and the latest HGConv~\cite{yu2020hybrid} show that HIRE is able to outperform its corresponding teacher model by 0.3\% $\sim$ 53.3\%. Overall, the improvements of our method are consistent and significant with better interpretability.

	The contributions of our work are summarized as follows:
	\begin{itemize}
		\item To the best of our knowledge, we are the first to investigate the knowledge distillation problem on heterogeneous graphs. Notably, we model the high-order knowledge of HGNNs by considering the second-order relational knowledge of heterogeneous graphs. 
		\item We propose a new distillation framework named HIRE, which focuses on individual node soft labels and correlations between different node types. Significantly, the latter transfers the relational knowledge between different nodes to the student by introducing a type-related attention layer to capture rich semantic information for different kinds of nodes in heterogeneous graphs.
		\item We conduct extensive experiments on popular HGNN models to evaluate the performance of the proposed framework. The results on ACM, IMDB, and DBLP datasets for node classification and node clustering exhibit consistently significant improvement, demonstrating our HIRE framework's effectiveness and generalization ability. 
	\end{itemize}

	The rest of this paper is organized as follows. In Section~\ref{sec:relatedwork}, we first revisit research works related to the studied problem. In Section~\ref{sec:method}, we introduce the framework and each component of the proposed method in detail. In Section~\ref{sec:experiment}, we analyze and evaluate the proposed framework through abundant experiments. Finally, conclusion is brought forth in Section~\ref{sec:conclusion}.

\section{Related Work}\label{sec:relatedwork}
	Since this paper focuses on combining HGNN with knowledge distillation to pursue higher performance, we discuss related works of HGNNs, and Knowledge Distillation in this section. In these areas, various approaches have been proposed over the past few years. We summarize them as follows.

\subsection{Heterogeneous Graph Neural Networks}
	GNNs~\cite{Scarselli2009TheGN} are initially proposed in 2009 and have achieved great success in various tasks dealing with non-grid data since the emergence of GCN~\cite{kipf2016semi}, such as node classification~\cite{oono2019graph}, link prediction~\cite{Zhang2018LinkPB}, and graph classification~\cite{errica2019fair}. GCN~\cite{kipf2016semi} simplifies the convolution operation through an efficient layer-wise propagation of node features. GAT~\cite{velivckovic2018graph} incorporates the attention mechanism into GCN to make it possible to learn the importance between nodes automatically. GraphSAGE~\cite{hamilton2017inductive} makes GCN scalable to a large-scale graph by sampling the neighbors. 
	However, the GNNs mentioned above can only handle homogeneous graphs. They cannot deal with heterogeneous graphs with various types of nodes and edges naturally. We refer the interested readers to~\cite{Wu2021ACS,Zhou2020GraphNN,zhang2020deep,Bacciu2020AGI,Chami2020MachineLO} for more comprehensive reviews on GNNs.

	Due to the ubiquity of heterogeneous graphs, researches have been witnessing tremendous growth in data mining and machine learning. They have been successfully applied in real-world systems such as recommendation~\cite{shi2018heterogeneous,zhao2017meta}, text analysis~\cite{linmei2019heterogeneous,kilimci2018deep}, and cybersecurity~\cite{liu2018heterogeneous,gutierrez2020multi}. The various HGNNs can be divided into two technical routes. One is based on meta-path. For example, HAN~\cite{wang2019heterogeneous} takes the first attempt to apply the attention mechanism to heterogeneous graphs, including node-level and semantic-level attentions. Considering the intermediate nodes in each meta-path that is ignored in HAN, MAGNN~\cite{fu2020magnn} proposes intra-meta-path aggregation and inter-meta-path aggregation methods. The other line of this research is based on edge type. These methods employ different sampling, aggregation functions for different types of neighbors and edges without utilizing meta-paths. To capture the heterogeneity of graphs, RGCN~\cite{schlichtkrull2018modeling} learns a special transformation matrix for each edge relation and embeds different nodes into different embedding spaces. HGT~\cite{hu2020heterogeneous} introduces the node-type and edge-type dependent attention mechanisms to handle graph heterogeneity. HGConv~\cite{yu2020hybrid} designs a hybrid convolution operation to model node representation by leveraging both node attributes and edge relation information.
	
	Nevertheless, these HGNN models still suffer from several limitations, as illustrated in Section~\ref{sec:introduction}. Hence, instead of designing a new HGNN, we focus on extracting the knowledge effectively inside different HGNN models.
	
\subsection{Knowledge Distillation}
	Knowledge distillation~\cite{hinton2015distilling} focuses on distilling the knowledge from a cumbersome model (i.e., teacher model) into a small model (i.e., student model), hoping the light-weight model can hold a similar performance as the teachers. Generally, there exist two categories of methods for knowledge distillation.  
	The first one is called logits-based distillation~\cite{hinton2015distilling,zhang2018deep,furlanello2018born}, aiming at matching the target output of the student model with the soft output distribution of the teacher model as much as possible.
	The other is termed as feature-based distillation~\cite{romero2014fitnets,komodakis2017paying,yim2017gift}, which uses the intermediate output to obtain a better student network. For example, Finet~\cite{romero2014fitnets} uses both the target outputs and the intermediate representations to extend further KD and adds a regressor on the intermediate layers to match the output of teachers and students of different sizes.

	In addition, there are some attempts to apply KD to GNNs. 
	LSP~\cite{Yang2020DistillingKF} utilizes a local structure-preserving method to measure the similarity of the local topological structures embedded by teachers and students. CPF~\cite{Yang2021ExtractTK} treats the student model as a trainable combination of parameterized label propagation and feature transformation modules to preserve structure-based and feature-based knowledge, respectively. GNN-SD~\cite{Chen2021OnSG} presents an ADR regularizer empowering the adaptive knowledge transfer inside a single GNN model by Self-distillation technology. Although these distillation methods achieve good performance, they are all designed for homogeneous graphs, where each node or edge is of the same type. 
	
	Heterogeneous graphs, such as movie networks containing different meaningful and complex semantic information, are ubiquitous in the real world but are ignored by existing knowledge distillation methods. Thereby, we design an effective distillation framework for heterogeneous graphs, named HIRE, to extract the knowledge of HGNNs both at node-level knowledge and relation-level knowledge. Our framework is very flexible and can be applied to an arbitrary HGNN architecture besides GCN.

\section{Method}\label{sec:method}
\begin{figure*}[h]
	\centering
	\includegraphics[width=0.8\textwidth]{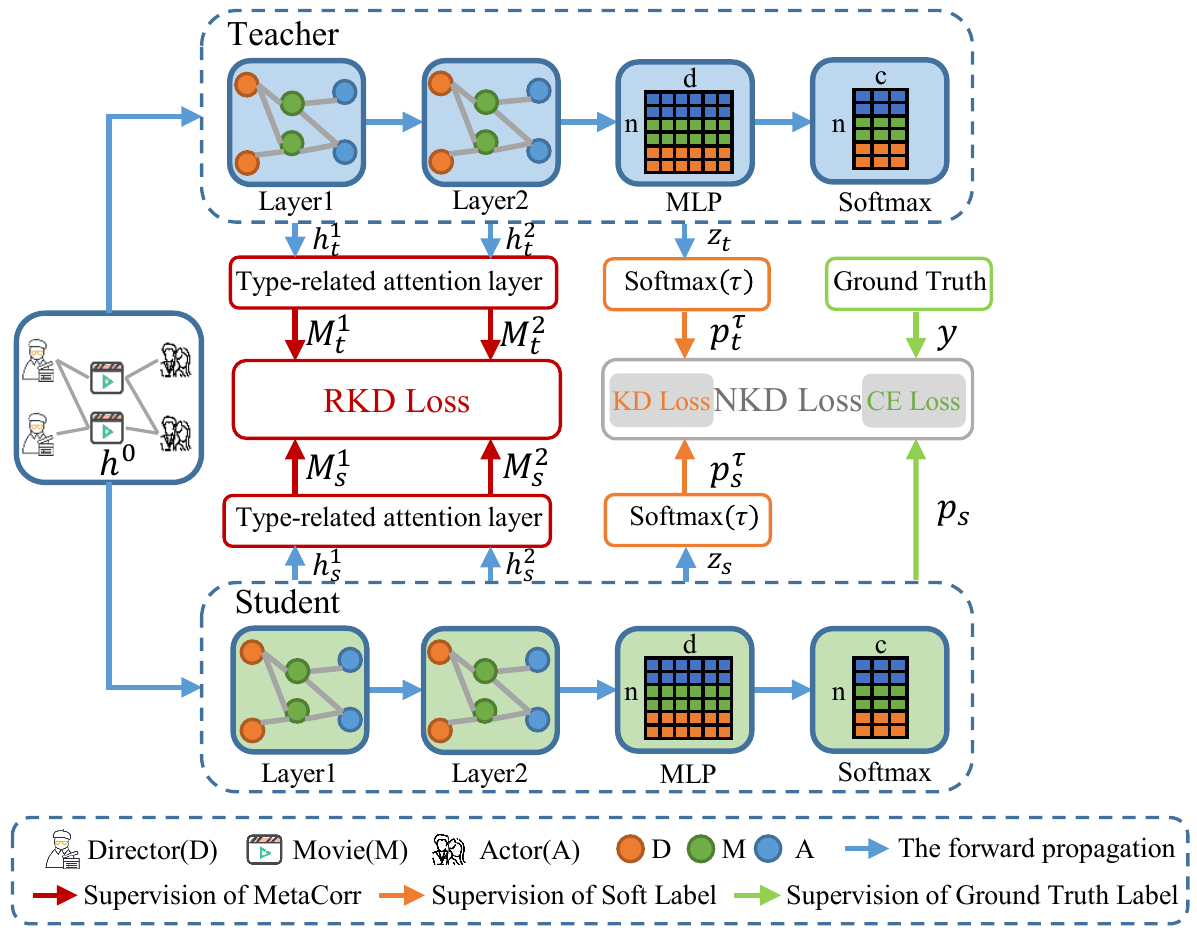}
    \caption{
    \textbf{HI}gh-order \textbf{RE}lational (HIRE) knowledge distillation framework. The student model is trained under the supervision of the correlation scores of RKD, soft labels of NKD, and ground truth labels transferred by the teacher model. Note that the model that provides knowledge is called Teacher, and the model that learns knowledge is called Student. The color indicates the node type. n, d, and c represent the number of nodes, the dimension of node features, and the number of classes, respectively.}
	\label{fig:framework}
\end{figure*}

In this section, we propose a novel \textbf{HI}gh-order \textbf{RE}lational (\textbf{HIRE}) knowledge distillation framework for HGNNs, which is depicted in Figure~\ref{fig:framework}. HIRE is constructed by two major components: node-level knowledge distillation (NKD) and relation-level knowledge distillation (RKD), where NKD aims to address the imprecise label problem and RKD is designed for modeling the semantic relation between different types of nodes in the heterogeneous graph. We first give a brief description of HGNNs and introduce the related notations. Table~\ref{tab:symbol} lists the meanings of our used mathematical symbols in this paper. Then we present our knowledge distillation framework to extract the knowledge of HGNNs both at node-level and relation-level. Finally, we give the overall loss and the overall training process of HIRE.

\subsection{Background and Notations}

\begin{table}[!t]
	\caption{Mathematical symbols and their meanings used in this paper.}
	\centering
	\begin{tabular}{l|p{0.8\linewidth}}
		\toprule
		Symbol & Meaning \\ 
		\hline
		$\alpha,\beta$ & the two hyperparamemters for balancing loss function\\
		\hline
		$\tau $ & the temperature parameter used in NKD \\
		\hline	
		$h_u$ & the hidden feature of node $u$ \\
		\hline
		$h_{t},h_{s}$ & a node feature representations learned by $T$ and $S$ \\
		\hline
		$p_{t},p_{s}$ & the final prediction probabilities of $T$ and $S$ \\
		\hline
		$q$ & the attention vector \\
		\hline
		$u,v, e$ & the target node $u \in \mathcal{V}$, the source node $v \in \mathcal{V}$, an edge $e \in \mathcal{E}$ in $\mathcal{G}$ \\
		\hline
		$x,y,N$ & the raw feature and one-hot label of a node, total number of nodes\\
		\hline
		$z_{t} ,z_{s} $ & the output logits of $T$ and $S$ \\
		\hline
		$\mathcal{D}$ & the training dataset with $N$ nodes defined as $\left \{ \left ( x_{i},y_{i}   \right )  \right \} _{i=1}^{N}$\\
		\hline
		$H_{t},H_{s}$ & the mean feature representations of the different node-set in $T$ and $S$\\
		\hline
		$\mathcal{N} _{u} $ & the set of neighbors of node $u$ \\
		\hline
		$\mathcal{N}_{u}^{r} $ & the set of neighbor indices of node $u$ under relation $r\in R$ \\
		\hline
		$R,r$ & the relation set, an edge relation $r \in R$ \\
		\hline
		$T,S $ & the teacher and student models\\
		\hline
		$\mathcal{V}, \mathcal{E}, \mathcal{G}$ & the vertex set, edge set, denotation of a graph \\
		\hline
		$W_t, W_s$ & the parameters of the teacher and the student \\
		\hline
		$\mathcal{Y}$ & the label set of nodes $\{y_1, y_2, \dots, y_N \}$, where $N=\vert \mathcal{V} \vert$ \\
		\bottomrule
	\end{tabular}
	\label{tab:symbol}
\end{table}

\textbf{Definition 1. Heterogeneous Graph~\cite{Sun2013MiningHI}.} A heterogeneous graph is defined as a graph $\mathcal{G} = \left ( \mathcal{V} ,\mathcal{E} , \mathcal{A} ,\mathcal{R } \right )$, in which $\mathcal{V} $ and $\mathcal{E} $ are the set of nodes and edges respectively. Each node $v\in \mathcal{V} $ and each edge $e\in \mathcal{E} $ are relevant to their mapping functions $\phi \left ( v \right ) :\mathcal{V} \to \mathcal{A} $ and $\varphi \left ( e \right ) :\mathcal{E} \to \mathcal{R} $. $\mathcal{A}$ and $\mathcal{R }$ denote the node types and edge types with the constraint of $\left |\mathcal{ A}  \right | + \left |\mathcal{ R}  \right |> 2  $.

\textbf{Definition 2. Relation.} A relation represents the interaction schema of the source node, the target node and the connected edge. Formally, for an edge $e= \left ( v,u \right ) $ with source node $v$ and target node $u$, the corresponding relation $r\in R$ is denoted as $e_{u,v}^{r}$. The inverse of $R$ is naturally represented by $R^{-}$, and we consider the inverse relation to propagate information of two nodes from each other. 

\textbf{Definition 3. Heterogeneous Graph Embedding~\cite{wang2022survey}.} Given a heterogeneous graph $\mathcal{G} = \left ( \mathcal{V} ,\mathcal{E} , \mathcal{A} ,\mathcal{R } \right )$, where nodes with type $A\in \mathcal{A} $ are associated with the attribute matrix
${X} _{A} \in \mathbb{R} ^{\left | \mathcal{V}_{A}   \right |\times d_{A}  } $, heterogeneous graph embedding aims to obtain the d-dimensional representation $h_{v} \in \mathbb{R} ^{d} $ for $v\in \mathcal{V} $, where $d \ll \left | V \right | $. The learned representations are able to capture both structural and semantic information, which could be applied in various tasks, such as node classification, node clustering and node visualization.

\subsection{Heterogeneous Graph Neural Networks}
In recent studies, the message-passing scheme is adopted by most recent GNNs~\cite{kipf2016semi,velivckovic2018graph,hamilton2017inductive}. Unlike the traditional GNNs that take homogeneous graphs as input, HGNNs can process heterogeneous graphs with more complex structure and attribute information. HGNN-based methods, along with related methods such as GCN, can be formalized as the following generic message-passing framework:
\begin{equation}
	h_{u}^{l} = U \left ( \underset{r\in R,v\in N_{u}^{r} }{A}  \left ( M_{r}  \left ( h_{u}^{l-1},h_{v}^{l-1},e_{v,u}^{r}    \right )  
	\right )  \right ) ,
	\label{eq:hgnn}
\end{equation} 
where $h_{u}^{l}$ denotes the node representation of node $u$ at layer $l$ by aggregating feature information from its local neighborhood $h_{v}$ along a certain edge feature $e_{v,u}^{r}$. $M_{r}$ is a message function defined on each edge relation $r$; $A$ is the aggregation function; $U$ is an update function defined on each node to update the feature of node $u$ by aggregating its incoming neighbor node $v$'s messages;

Similar to the GNNs, the key of HGNNs is how to design the appropriate aggregation function to capture the semantics contained in the neighborhood, i.e., $A$ in Eq~\ref{eq:hgnn}. Some popular HGNNs' message-passing approaches are summarized in Table~\ref{tab:hgnn-agg}. These methods are mainly divided into metapath-based and edge-based HGNN models. After that, a large number of metapath-based HGNNs have emerged, such as MAGNN~\cite{fu2020magnn}, which further considers the intra-metapath aggregation to incorporate intermediate nodes along the meta-path on the basis of HAN. Another model, PSHGAN~\cite{mei2022heterogeneous}, introduces meta-structure~\cite{Huang2016MetaSC} based on HAN.

\begin{table}[htbp]
	\caption{A summary of message-passing-based HGNN models. Additional notations: $\phi _{v}$ is a linear function mapping $v$'s vector back to its node type-specific distribution; $\sigma$ denotes an activation function, such as the $ReLU\left ( \cdot  \right ) = max\left ( 0,\cdot  \right ) $; $W$ is a layer-specific trainable weight matrix; $m$ is one meta-path and $P$ is the total number of meta-paths (for more details, see~\cite{wang2019heterogeneous}).
	}
	\centering
	\resizebox{0.5\textwidth}{!}{
		\begin{tabular}{c|c}
			\toprule 
			Models & Aggregation Function \\ \hline
			RGCN~\cite{schlichtkrull2018modeling}   &  $h_{u}^{l} =\sigma \left ( \sum_{r\in R }^{} \sum_{v\in N_{u}^{r} }^{} \frac{1}{\left | N_{u}^{r} \right |}W_{r}^{l-1}h_{v}^{l-1}+  W_{0}^{l-1}h_{u}^{l-1}  \right )  
			$                   \\ \hline
			HAN~\cite{wang2019heterogeneous}    &  $h_v^{m} =\sigma \left ( \sum_{u\in N_{v}^{m}  }^{} NodeAttention_{u,v}^{m} \cdot h_{u}\right ) ;
			h_{v}= \sum_{m }^{P}SemanticAttention^{m} \cdot h_{v}^{m} 
			$			        \\ \hline
			HGT~\cite{hu2020heterogeneous}    &  $\hat{h} _{v}^{l} =\sum_{u\in N_{v}^{r} }^{} Attention\left ( u,v \right ) \odot Message\left ( u,v,e_{vu} \right );
			h_{v}^{l}  =\phi _{v} \left ( \sigma \left ( \hat{h} _{v}^{l} \right )  \right ) +h_{v}^{l}   $                     \\ \hline
			HGConv~\cite{yu2020hybrid} &  $h_{v,r}^{l}=\sigma \left ( \sum_{u\in N_{v}^{r} }^{} MicroAttention\cdot h_{v}^{l-1}  \right ) 
			;  
			\hat{h}_{v}^{l} =\sum_{r\in R }^{} MacroAttention\cdot h_{v,r}^{l}  
			$                     \\ \bottomrule 
		\end{tabular}
	}
	\label{tab:hgnn-agg}
\end{table}

\subsection{Node-level Knowledge Distillation}
As we all know, a target node's label plays a vital role in the downstream task, such as movie classification in IMDB. However, as mentioned in Section~\ref{sec:introduction}, the data labelings of target nodes are imprecise. The node-level knowledge distillation is applied
to transfer soft labels ($p_{t_{i}}^{\tau }$) 
predicted by the teacher network to the student network, allowing students to learn smoother knowledge~\cite{hinton2015distilling}. The soft labels can be obtained using softmax with temperature scaling, providing more inter-class and intra-class information. The distillation loss for soft logits can be denoted as:
\begin{equation}
	\begin{split}
		\mathcal{L}_{KD} &= \frac{1}{N} \sum_{i=1}^{N} \tau^{2}  D\left ( p_{s_{i}}^{\tau },p_{t_{i}}^{\tau }  \right )  	,\\
        p_{i}^{\tau }  &=\frac{exp\left ( z_{i}/\tau  \right ) }{{\sum_{j=1}^{C}}exp\left ( z_{j}/\tau \right ) } ,\\
	\end{split}
	\label{eq:kdloss1}
\end{equation}
where $D\left ( \cdot  \right ) $ is the distance function (e.g., Kullback-Leibler divergence) to minimize the difference between the logits produced by the teacher model and those produced by the student model. $z_{i} $ is the corresponding logit of the $i$-th class. $\tau$ is a temperature hyperparameter intended to smooth outputs from teacher models. As shown in Figure~\ref{fig:framework}, the teacher and student models use the softmax function with $\tau$ after their respective MLP layer to obtain $p_{t_{i}}^{\tau }$ and $p_{s_{i}}^{\tau }$. A larger $\tau$ leads to a smoother probability distribution over classes, promoting the student model to learn more smoothing information about which classes are more similar to the predicted class.
. 

Besides the teacher's soft labels, it can further boost the training performance with existing ground truth labels, as stated by Hinton et al.~\cite{hinton2015distilling}, which can significantly improve the performance of the student model. Accordingly, the NKD loss can be written as a weighted average of the cross-entropy loss and the distillation loss:
\begin{equation}
	\mathcal{L}_{NKD} = \left ( 1-\alpha  \right ) \mathcal{L} _{CE} +  \alpha \mathcal{L} _{KD}, 
	\label{eq:kdloss}
\end{equation}
where $\mathcal{L} _{CE}= - \sum_{i}^{N} y_{i} log\left ( p_{s_{i}}  \right ) $ is the basic cross-entropy loss, $i$ denotes the $i$-th node, $\alpha$ is the hyperparamemter for balancing the cross-entropy loss and the distillation loss.

Nevertheless, NKD cannot address the semantic relation modeling issue
because it ignores the crucial correlation between different types of nodes in heterogeneous graphs, which can be solved in the next section.

\subsection{Relation-level Knowledge Distillation}
Generally, each node in a heterogeneous graph contains multiple types of semantic relations information. 
To learn a more comprehensive node embedding, the various semantics revealed by various edge relations need to be considered. 
Intuitively, we observe that the correlation between different node embeddings of the intermediate convolutional layer of the teacher model is also essential for downstream tasks. It directly reflects how the teacher models high-order relations between different nodes in the embedded feature space.

The relation-level knowledge distillation is further proposed to distill the dark knowledge hidden in different relations of HGNNs. More specifically, a MetaCorr correlation matrix is designed to encode relation-level knowledge between different types of nodes from pre-trained HGNNs, and a type-related attention layer is introduced to learn the importance of different node types automatically.

Firstly, let $H_{t}$ be the feature representations of different node types in the $l$-th layer of the teacher. It can be obtained by performing mean operation on $h_{t}$:
\begin{equation}
    	H_{t} =\left \{ \bar{h}_{t}^{1}, \dots  ,\bar{h}_{t}^{k},\dots , \bar{h}_{t}^{K} \right \}  
    	\label{eq:feats-t}
\end{equation}
where $\bar{h}_t^k = mean\left ( \left \{ h_t^i \right \} \right ), i \in \mathcal{V}^k $, $k$ is a node type, $\mathcal{V}^k$ is the node set of the type $k$, $h$ is the node embedding, and $K$ is the number of node types. Similarly, in this way we can also get $H_{s}$.

To encode the relation-level knowledge, a correlation matrix is further designed
to help students imitate the relational correlation between node embeddings of different types generated by the teacher HGNN, called $\mathbf{MetaCorr} $, which is defined as:
\begin{equation}
	MetaCorr_{t}^{i,j} = \phi \left ( H_{t}^{i} ,H_{t}^{j} \right ) 	
	\label{eq:metacorr-t}
\end{equation}
where $i$, $j$ represent different types of nodes. $\phi\left ( x,y \right )  = e^{-\frac{1}{2\sigma ^{2} } {\left \| x -y \right \|} ^2 }   $ is to measure the similarity between two node embeddings, where a larger $\phi$ indicates that two-node embeddings are farther apart. The reason for the usage of Gaussian RBF kernel~\cite{Peng2019CorrelationCF} is that Gaussian RBF is more flexible and powerful in capturing the complex non-linear relationship between various nodes. The second-order Taylor expansion on $\phi$ is chosen to avoid dimensional disaster. In the same way, we can also get $MetaCorr_{s}^{i,j} = \phi \left ( H_{s}^{i} ,H_{s}^{j} \right )$.

Considering that different node embeddings in heterogeneous graphs play different roles for downstream tasks such as classification, an attention mechanism is used to learn the most relevant node adaptively to make decisions. 
As depicted in Figure~\ref{fig:example}, taking the movie classification task on IMDB as an example, it is obvious that Movie plays a leading role in determining the type of movie among Movie, Actor, and Director nodes, which is also verified in Section~\ref{sec:rq2_att}. Here we focus on type $i$ in $S$, whose embedding in $H_{s}$ is $ H_{s}^{i}$. First, perform a nonlinear transformation to transform the embedding and then apply one shared attention vector $q$ to get the attention value $\omega_{s}^{i}$ as follows:
\begin{equation}
	\omega_{s}^{i}=q^{T} \cdot  tanh\left ( W_{s} \cdot \left ( H_{s}^{i}  \right ) ^{T} +  b_{s}  \right )   ,
	\label{eq:att}
\end{equation}
where $W_{s}$ is the weight matrix, $b$ is the bias vector, $q$ is the shared attention vector. Similarly, we can get the attention values $\omega_{t}^{i}$ of type $i$ based on  $H_{t}^{i}$. Then normalize the attention values $w_{s}^{i}$ to get the final attention coefficients via softmax function:
\begin{equation}
	\alpha_{s}^{i}=softmax\left ( \omega_{s}^{i}   \right ) =\frac{exp\left ( \omega_{s}^{i}  \right ) }{  {\textstyle \sum_{j=1}^{k}exp\left ( \omega_{s}^{j}  \right ) } }  .
	\label{eq:att-nodes}
\end{equation}
Obviously, the higher $\alpha_{s}^{i}$ represents a more critical node, and $\alpha ^{i}$ can be adjusted dynamically during model training. The RKD loss can be denoted as:
\begin{equation}
	\mathcal{L}_{RKD} =\sum_{i=1}^{K} \sum_{j=1}^{K} \alpha _{i} D\left (  MetaCorr_{s}^{i,j},MetaCorr_{t}^{i,j} \right )  ,
	\label{eq:metacorr-loss}
\end{equation}
where $D$ is the mean square error for distance metric here.

\subsection{The Overall Loss}
Finally, based on the above components (i.e., NKD and RKD), HIRE can successfully extract the high-order relational knowledge from pre-trained HGNN. The total loss of HIRE for training the student model can be denoted as:
\begin{equation}
	\mathcal{L} =  \mathcal{L} _{NKD}  +  \beta \mathcal{L}_{RKD} , 
	\label{eq:totalloss}
\end{equation}
where $\beta$ is the hyperparameter for balancing the first-order Node-level Knowledge Distillation and the second-order Relation-level Knowledge Distillation. 

HIRE can be optimized in an end-to-end manner. The overall training process of HIRE is shown in Algorithm~\ref{alg:algorithm1}. In the total loss function in Eq.~\ref{eq:totalloss}, 
use grid search to select the best hyperparameters, including $\tau$ in $\left [ 1,2,\cdots ,10  \right ]$, $\alpha$ in $\left [0, 0.1,\cdots ,1  \right ] $, and $\beta $ in $\left [ 0.01,0.1,1,10,100  \right ] $. The sensitivity analysis of the three parameters can be found in Section~\ref{sec:rq3}.

\subsection{Training}
We refer to the pre-trained teacher HGNN with fixed parameter ${W}_{t} $ as $T$ and the same model configuration with the training parameter $W_{s} $ as $S$. 
The intermediate convolution layer's feature representations learned by $T$ and $S$ are respectively denoted as $h_{t} $ and $h_{s}$. Similarly, $z_{t} $ and $z_{s}$ are the output logits of the teacher and student networks. $p_{t}=\delta\left ( z_{t}  \right ) $ and $p_{s}=\delta\left ( z_{s}  \right )$ represent the final prediction probabilities of teacher and student respectively, where $\delta \left ( \cdot  \right ) $ denotes the softmax function.
\begin{algorithm}[t]
	\caption{HGNN Knowledge Distillation.}
	\label{alg:algorithm1}
	\KwIn{Training dataset $\mathcal{D} = \left \{ \left ( x_{i},y_{i}   \right )  \right \} _{i=1}^{N} $; A pre-trained teacher model with fixed parameter $W_{t} $; A student model with random initialized parameters $W_{s} $ .}
	\KwOut{A well-trained student model.}  
	\BlankLine
	Initialize $W_{s}$ randomly;
	
	\While{\textnormal{not converged}}{
		Put full batch data to $T$ and $S$;
		
		\ForEach{each sample in $\mathcal{D}$}{
			Obtain node representation of the intermediate convolution layer $h_{t} $, $h_{s} $ and prediction of the linear transformation layer $p_{t}$ , $p_{s}$;
			
			Evaluate $\mathcal{L}_{NKD}$ using Eq.~\ref{eq:kdloss};
			
			Compute adapted attention coefficients $\alpha _{s}^{i} $ using Eq.~\ref{eq:att-nodes};
			
			Evaluate $\mathcal{L}_{RKD}$ using Eq.~\ref{eq:metacorr-loss};
		}
		Update $W_{s}$ by minimize the total loss in Eq.~\ref{eq:totalloss}; 
		
	}
	
	Finally obtain the optimized parameters $W_{s} $.

\end{algorithm}

\textbf{Algorithm.} Algorithm~\ref{alg:algorithm1} outlines the model training process of HIRE. 
Before proceeding with the distillation, we first pre-train a teacher model using fixed parameters $W_{t}$ in line 1. In line 5, the intermediate node representations $h_{t}$, $h_{s}$  are retrieved for RKD, and the prediction logits of each model $p_{t}$, $p_{s}$ are extracted for NKD. Line 6 depicts the soft label transfer from the teacher model to the student model following Eq.~\ref{eq:kdloss} by using $p_{t}$, $p_{s}$.
Line 7-8 show the attentive relational knowledge transfer of RKD follwing Eq.~\ref{eq:att-nodes} and Eq.~\ref{eq:metacorr-loss} by using $h_{t}$ and $h_{s}$. In general, we utilize Eq.~\ref{eq:totalloss} in line 10 to minimize the total loss, which considers both node-level and relation-level knowledge distillation. Finally, we get a well-trained student model with the optimized parameters $W_{s}$.

\textbf{Discussion.} Since the proposed HIRE is independent of the model architectures of HGNNs, it can be widely applied to different HGNN models, including RGCN, HGT, and HGConv. Compared with the previous distillation methods on homogeneous graphs, HIRE can integrate rich semantics for heterogeneous graphs. Benefiting from RKD, different node representations can enhance and promote each other, whose dynamic attention coefficients also partially explain the distillation process. Based on the attention value of the type-related attention layer, we can observe which node makes higher contributions to our downstream tasks, such as node classification, node clustering, and node visualization. The experimental results can be found in Section~\ref{sec:experiment}. 
Compared to the adopted deep learning classifiers in~\cite{Aydn2020DeepLC}, our proposed HIRE presents an end-to-end paradigm for the complex network classification, which alleviates the feature engineering efforts like the pre-processing PCA method. It could tackle the network classification problem since it could mine the graph structure and connectivity among brain regions, and refine the node representation. More GNN-based methods have also been proposed to deal with the complex brain network~\cite{Yan2022ModelingSP,bessadok2021graph,cui2021brainnnexplainer,bi2020gnea}.

\section{Experiment}\label{sec:experiment}
In this section, 
elaborate experiments are conducted to demonstrate the efficacy of HIRE for various HGNNs.
We aim to address the following key research questions:

\begin{itemize}
    \item \textbf{RQ1:} How does our HIRE perform with various HGNNs on three tasks: node classification, node clustering, and node visualization?
    \item \textbf{RQ2:} What are the impacts of the crucial second-order relation-level knowledge distillation in our HIRE framework?
    \item \textbf{RQ3:} How do different settings of hyper-parameters influence the prediction performance of our HIRE framework?
\end{itemize}

\subsection{Datasets}
Experiments are conducted
on three widely used real-world heterogeneous graph datasets. Statistics of the three datasets are summarized in Table~\ref{tab:data}. 

\begin{table}[htbp]
	\caption{Statistics of the heterogeneous graph datasets used for training and evaluations. \# represents the number of nodes or edges.}
	\centering 
	\resizebox{1.0\linewidth}{!}{ 
	
	\begin{tabular}{l|c|c|c|c}
		\toprule 
		Dataset&Node&Edge&Meta-path&Data Split\\ 
		\midrule 
		\multirow{3}{*}{ACM} & \#Paper(P): 4025 & \#P-A: 13407 & PAP & Train: 808\\
		& \#Author(A): 17351 & \#P-F: 4025&PFP&Validation: 401\\
		& \#Field(F): 72&&&Test: 2816\\
		
		\midrule 
		\multirow{3}{*}{IMDB} & \#Movie(M): 4278 & \#M-D: 4278 & MDM & Train: 858\\
		& \#Director(D): 2081 & \#M-A: 12828&MAM&Validation: 427\\
		& \#Actor(A): 5257&&&Test: 2993\\
	
		\midrule 
		\multirow{4}{*}{DBLP} & \#Author(A): 4057 & \#P-A: 19645 & APA & Train: 800\\
		& \#Paper(P): 14328 & \#P-T: 85810&APTPA&Validation: 400\\
		& \#Term(T): 7723 & \#P-C: 14328&APCPA&Test: 2857\\
		& \#Conf(C): 20&&&\\
		\bottomrule  
	\end{tabular}
	}
	\label{tab:data}

\end{table}

\begin{itemize}
	\item \textbf{ACM\footnote{http://dl.acm.org/}}: provided by~\cite{wang2019heterogeneous}, is an academic network dataset, containing 4025 papers (P), 17351 authors (A), and 72 fields (F). By default, 
	the author nodes are split into 
	training, validation, and testing sets of 808 (20.08\%), 401 (9.96\%), and 2816 (69.96\%) nodes, respectively. 
	
	\item \textbf{IMDB\footnote{https://www.imdb.com/}}: provided by~\cite{fu2020magnn}, is a movie network dataset, including 4278 movies (M), 2081 directors (D) and 5257 actors (A). By default, the movie nodes are split into training, validation, and testing sets of 858 (20.06\%), 427 (9.98\%), and 2993 (69.96\%) nodes, respectively.

	\item \textbf{DBLP\footnote{https://dblp.uni-trier.de}}: provided by~\cite{fu2020magnn,mei2022heterogeneous}, is an academic network dataset, consisting of 4057 authors (A), 14328 papers (P), 7723 terms (T), and 20 conferences (C). By default, the author nodes are split into training, validation, and testing sets of 800 (19.72\%), 400 (9.86\%), and 2857 (70.42\%) nodes, respectively.

\end{itemize}

\subsection{Experiment Setting} \label{sec:experimentsetting}

\subsubsection{Teacher Methods}
For a comprehensive comparison, 
six models in the Teacher-Student knowledge distillation framework of HIRE are considered,
including two homogeneous graph models and four heterogeneous graph models.
\begin{itemize}
	\item 
	\textbf{GCN}~\cite{kipf2016semi} performs convolutional operations via a localized first-order approximation of spectral graph convolutions. 
	All meta-paths for GCN are tested and the best performance is reported from the best meta-path.
	
	\item 
	\textbf{GAT}~\cite{velivckovic2018graph} introduces the attention mechanism into GCN and assigns different importance to different nodes. 
	All meta-paths for GAT are tested and the best performance is reported from the best meta-path in the same way as GCN.

	\item 
	\textbf{RGCN}~\cite{schlichtkrull2018modeling} directly leverages different weight matrices to model different edges of the original heterogeneous graph and introduces two kinds of regularizations to reduce the number of parameters for modeling amounts of edge relations. 
	
	\item \textbf{HAN}~\cite{wang2019heterogeneous} exploits the hierarchical attention mechanism, including node-level attention and semantic-level attention, to learn the importance of nodes and meta-paths simultaneously.

	\item \textbf{HGT}~\cite{hu2020heterogeneous} introduces type-specific transformation matrices to learn the importance of different nodes by using Transformer architecture ~\cite{vaswani2017attention}.
	
	\item \textbf{HGConv}~\cite{yu2020hybrid} introduces a hybrid micro/macro level convolution to learn node representations on heterogeneous graphs. The former convolution can learn the importance of nodes within the same relation, and the latter one can distinguish the subtle difference across different relations.
\end{itemize}
For the fairness of the experiment, we set all models' layers as 2.
Please refer to~\ref{sec:appendixA} in the appendix for more details on the experimental setting.

\subsubsection{Experimental Setup}
\textbf{Baseline Settings.} Following~\cite{wang2019heterogeneous}, we apply PAP and PSP on ACM, and select MDM and MAM on IMDB. 
Similarly, APA, APTPA, and APCPA are used as meta-paths for DBLP. For GNNs, including GCN and GAT, we test all the meta-paths mentioned above and find that their best meta-path on ACM, IMDB, and DBLP are PAP, MDM, and APCPA, respectively. For HAN, all of the above meta-paths are used. For edge-based HGNNs, including RGCN, HGT, and HGConv, all the edges and their reverse edges mentioned in Table~\ref{tab:data} are used to construct heterogeneous graphs. All teacher models and student models are initialized with the same parameters suggested by their papers, and we further carefully tune parameters to get students' optimal performance. 
By default, the datasets are split into 
the training set, validation set, and test set according to the same ratio of 2:1:7 and further divide the training set into 20\%, 40\%, 60\%, and 80\% ratio. 

\textbf{Experimental Settings.}
To  present experimental details completely, we point out: 
1)The Adam~\cite{Kingma2015AdamAM} optimizer is employed
 except for AdamW used in HGT. 
2) The grid search is adopted to select the best hyperparameters, including dropout~\cite{Srivastava2014DropoutAS} in $\left [ 0.0,0.1\cdots,1.0 \right ] $, learning rate and weight decay in $\left [ 0.0001,0.001,0.01,0.1 \right ]$. 3) Three hyper-parameters in Eq.~\ref{eq:totalloss} are required to train students guided by HIRE, where we set $\alpha \in \left [ 0.0,0.1,\cdots,1.0  \right ] $, $\beta  \in \left [ 0.01,0.1,1,10,100  \right ] 
$, $\tau \in \left [ 1,2,\cdots ,10  \right ] $. HIRE is not sensitive to these hyper-parameters, which will be discussed in Section~\ref{sec:rq3}. 
4)All the teacher and student models are trained
with a fixed 200 epochs and are reported the average and standard deviation results for all models by running five times with different seeds $\in \left \{ 0,1,2,3,4 \right \}.$ 
5) For RGCN, HGT, and HGConv, different types of nodes remain in hidden representations of each convolution layer, which are well suitable for our HIRE model. However, for homogeneous GNNs (i.e., GCN and GAT) and metapath-based HGNNs (i.e., HAN), our RKD degenerates into a constant in the loss function because of the homogeneous node representations in hidden convolution layers.

All experiments conducted in this paper run on a NVIDIA Tesla V100 GPU with 32 GB memory, which are implemented by PyTorch\footnote{https://pytorch.org/}~\cite{Paszke2019PyTorchAI} of version 1.6.0 and Deep Graph Library (DGL)\footnote{https://www.dgl.ai/}~\cite{wang2019deep} of version 0.6.1.

\subsubsection{Evaluation Tasks and Metrics}
We perform three tasks on ACM, IMDB, and DBLP datasets: node classification, node clustering, and node visualization.

Firstly, We perform node classification on the three datasets to compare the performance of different models. To enrich the experiment, the ratio of training data is set to $20\%\sim 80\%$. We use Micro-F1 and Macro-F1 evaluation metrics on the node classification task. Experimental
results are shown in Table~\ref{tab:clf-acm}, Table~\ref{tab:clf-imdb} and Table~\ref{tab:clf-dblp}, which are reported as the mean and variance of classification metrics with five runs..

We also perform clustering tasks to evaluate the node embeddings learned from the HIRE framework. We first obtain the node representation by feeding forward on the trained model and then feed the normalized node representation into the K-Means algorithm. We set the number of clusters in each dataset to the number of actual classes (i.e., ACM, IMDB, and DBLP are 3, 3, and 4, respectively). ARI and NMI are adopted as the evaluation metrics to assess the quality of clustering results, as seen in Table~\ref{tab:clu}.

In addition to the quantitative evaluation of the node embedding, we also visualize the embedding of the target node and evaluate its embedding result qualitatively. Specifically, we utilize t-SNE~\cite{Maaten2008VisualizingDU} to project the learned embeddings of each HGNN constructed with HIRE into a 2-dimensional space. Based on the learned node visualization, we can clearly observe the improvements of existing HGNNs guided by our HIRE.

\subsection{RQ1: Overall performance}

\subsubsection{Node Classification}
As drawn in Table~\ref{tab:clf-acm}, Table~\ref{tab:clf-imdb} and Table~\ref{tab:clf-dblp}, HIRE has achieved consistent and significant performance improvement over their teacher model as the training ratio increases, demonstrating the effectiveness and generalization ability of our HIRE framework.  

\begin{table*}[htbp]

\caption{HIRE improves various teacher models with satisfactory margins on ACM. We report the Micro-F1 and Macro-F1 on ACM for the node classification task. Teacher indicates the pre-trained model with a basic cross-entropy loss. All the experiments are run five times with different random seeds in each row. Bold font reports the best performance. The superscript represents the standard deviation. Table~\ref{tab:clf-imdb} and Table~\ref{tab:clf-dblp} follow the same experiment settings as this table.
	}
\centering
\resizebox{0.9\textwidth}{!}{	
	
\begin{tabular}{|c|c|c|c|c|c|c|c|c|c|}
\hline
\textbf{Datasets}     & \textbf{Metrics}           & \textbf{Training}    & \textbf{Network} & \textbf{GCN} & \textbf{GAT} & \textbf{RGCN}         & \textbf{HAN}          & \textbf{HGT}          & \textbf{HGConv}       \\ \hline
\multirow{24}{*}{ACM} & \multirow{12}{*}{Micro-F1} & \multirow{3}{*}{0.2} & Teacher          & 88.02 $^{ 0.65\rule{0pt}{0.2cm}}$ & 89.42 $^{ 0.55}$ & 89.86 $^{ 1.12}$          & 90.50 $^{ 0.58}$          & \textbf{90.97 $^{ 0.56}$} & 90.72 $^{ 0.89}$          \\ \cline{4-10} 
                      &                            &                      & HIRE             & 89.00 $^{ 0.46\rule{0pt}{0.2cm}}$ & 90.19 $^{ 0.50}$ & \textbf{92.56 $^{ 0.60}$} & 91.73 $^{ 0.40}$          & 91.43 $^{ 0.35}$          & 92.20 $^{ 0.32}$          \\ \cline{4-10} 
                      &                            &                      & Improve          & 1.1\%        & 0.9\%        & \textbf{3.0\%}        & 1.4\%                 & 0.5\%                 & 1.6\%                 \\ \cline{3-10} 
                      &                            & \multirow{3}{*}{0.4} & Teacher          & 90.13 $^{ 0.65\rule{0pt}{0.2cm}}$ & 90.85 $^{ 0.78}$ & 91.63 $^{ 0.47}$         & \textbf{93.68 $^{ 0.17}$} & 92.56 $^{ 0.33}$          & 93.06 $^{ 0.53}$          \\ \cline{4-10} 
                      &                            &                      & HIRE             & 90.77 $^{ 0.31\rule{0pt}{0.2cm}}$ & 91.34 $^{ 0.40}$ & 93.64 $^{ 0.16}$         & \textbf{93.89 $^{ 0.18}$} & 93.17 $^{ 0.24}$          & 93.71 $^{ 0.23}$          \\ \cline{4-10} 
                      &                            &                      & Improve          & 0.7\%        & 0.5\%        & \textbf{2.2\%}        & 0.2\%                 & 0.7\%                 & 0.7\%                 \\ \cline{3-10} 
                      &                            & \multirow{3}{*}{0.6} & Teacher          & 92.23 $^{ 0.47\rule{0pt}{0.2cm}}$ & 92.19 $^{ 0.19}$ & 94.37 $^{ 0.18}$         & 95.38 $^{ 0.70 }$          & 95.16 $^{ 0.28}$          & \textbf{95.73 $^{ 0.18}$} \\ \cline{4-10} 
                      &                            &                      & HIRE             & 92.64 $^{ 0.30\rule{0pt}{0.2cm}}$ & 92.29 $^{ 0.28}$ & 95.63 $^{ 0.22}$          & 95.59 $^{ 0.26}$          & 95.59 $^{ 0.27}$          & \textbf{95.97 $^{ 0.25}$} \\ \cline{4-10} 
                      &                            &                      & Improve          & 0.4\%        & 0.1\%        & \textbf{1.3\%}        & 0.2\%                 & 0.5\%                 & 0.3\%                 \\ \cline{3-10} 
                      &                            & \multirow{3}{*}{0.8} & Teacher          & 93.46 $^{ 0.61\rule{0pt}{0.2cm}}$ & 93.01 $^{ 0.23}$ & 97.19 $^{ 0.27}$         & 96.58 $^{ 0.41 }$         & \textbf{97.46 $^{ 0.27}$} & 96.78 $^{ 2.02}$          \\ \cline{4-10} 
                      &                            &                      & HIRE             & 93.79 $^{ 0.20\rule{0pt}{0.2cm}}$ & 92.96 $^{ 0.26}$ & 97.79 $^{ 0.06}$          & 96.75 $^{ 0.25}$          & 97.78 $^{ 0.21}$          & \textbf{98.03 $^{ 0.19}$} \\ \cline{4-10} 
                      &                            &                      & Improve          & 0.4\%        & 0.0\%        & 0.6\%                 & 0.2\%                 & 0.3\%                 & \textbf{1.3\%}        \\ \cline{2-10} 
                      & \multirow{12}{*}{Macro-F1} & \multirow{3}{*}{0.2} & Teacher          & 87.83 $^{ 0.80\rule{0pt}{0.2cm}}$ & 89.38 $^{ 0.62}$ & 89.83 $^{ 1.24}$          & 90.37 $^{ 0.68}$          & \textbf{90.85 $^{ 0.64}$} & 90.62 $^{ 0.92}$          \\ \cline{4-10} 
                      &                            &                      & HIRE             & 88.96 $^{ 0.41\rule{0pt}{0.2cm}}$ & 90.13 $^{ 0.57}$ & \textbf{92.59 $^{ 0.65}$} & 91.73 $^{ 0.46}$          & 91.36 $^{ 0.54}$          & 92.19 $^{ 0.39}$          \\ \cline{4-10} 
                      &                            &                      & Improve          & 1.3\%        & 0.8\%        & \textbf{3.1\%}        & 1.5\%                 & 0.6\%                 & 1.7\%                 \\ \cline{3-10} 
                      &                            & \multirow{3}{*}{0.4} & Teacher          & 89.94 $^{ 0.78\rule{0pt}{0.2cm}}$ & 90.79 $^{ 0.88}$ & 91.63 $^{ 0.55}$          & \textbf{93.71 $^{ 0.18}$} & 92.49 $^{ 0.37}$          & 93.09 $^{ 0.50}$          \\ \cline{4-10} 
                      &                            &                      & HIRE             & 90.75 $^{ 0.31\rule{0pt}{0.2cm}}$ & 91.35 $^{ 0.40}$ & 93.64 $^{ 0.16}$          & \textbf{93.93 $^{ 0.20}$} & 93.13 $^{ 0.26}$          & 93.73 $^{ 0.27}$          \\ \cline{4-10} 
                      &                            &                      & Improve          & 0.9\%        & 0.6\%        & \textbf{2.2\%}        & 0.2\%                 & 0.7\%                 & 0.7\%                 \\ \cline{3-10} 
                      &                            & \multirow{3}{*}{0.6} & Teacher          & 92.17 $^{ 0.54\rule{0pt}{0.2cm}}$ & 92.22 $^{ 0.21}$ & 94.36 $^{ 0.20}$          & 95.41 $^{ 0.67}$          & 95.11 $^{ 0.32}$          & \textbf{95.72 $^{ 0.18}$} \\ \cline{4-10} 
                      &                            &                      & HIRE             & 92.67 $^{ 0.32\rule{0pt}{0.2cm}}$ & 92.33 $^{ 0.27}$ & 95.65 $^{ 0.25}$          & 95.60 $^{ 0.28}$          & 95.54 $^{ 0.30}$          & \textbf{95.95 $^{ 0.27}$} \\ \cline{4-10} 
                      &                            &                      & Improve          & 0.5\%        & 0.1\%        & \textbf{1.4\%}        & 0.2\%                 & 0.5\%                 & 0.2\%                 \\ \cline{3-10} 
                      &                            & \multirow{3}{*}{0.8} & Teacher          & 93.42 $^{ 0.68\rule{0pt}{0.2cm}}$ & 93.08 $^{ 0.25}$ & \textbf{97.20 $^{ 0.27}$} & 96.62 $^{ 0.40}$          & 97.45 $^{ 0.29}$          & 96.84 $^{ 1.87}$          \\ \cline{4-10} 
                      &                            &                      & HIRE             & 93.85 $^{ 0.19\rule{0pt}{0.2cm}}$ & 93.04 $^{ 0.27}$ & 97.79 $^{ 0.05}$          & 96.76 $^{ 0.28  }$        & 97.78 $^{ 0.22}$          & \textbf{98.04 $^{ 0.20}$} \\ \cline{4-10} 
                      &                            &                      & Improve          & 0.5\%        & 0.0\%        & 0.6\%                 & 0.1\%                 & 0.3\%                 & \textbf{1.2\%}        \\ \hline
\end{tabular}

}
\label{tab:clf-acm}
\end{table*}

\begin{table*}[htbp]
\caption{HIRE improves teacher models baselines with satisfactory margins  on IMDB. }
	
\centering
\resizebox{0.9\textwidth}{!}{

\begin{tabular}{|c|c|c|c|c|c|c|c|c|c|}
\hline
\textbf{Datasets}      & \textbf{Metrics}           & \textbf{Training}    & \textbf{Network} & \textbf{GCN} & \textbf{GAT} & \textbf{RGCN}         & \textbf{HAN}          & \textbf{HGT}    & \textbf{HGConv} \\ \hline
\multirow{24}{*}{IMDB} & \multirow{12}{*}{Micro-F1} & \multirow{3}{*}{0.2} & Teacher          & 55.86 $^{ 0.43\rule{0pt}{0.2cm}}$ & 56.67 $^{ 0.25}$ & 53.24 $^{ 0.53 }$         & \textbf{60.74 $^{ 0.34}$} & 57.16 $^{ 0.19}$    & 56.43 $^{ 0.42}$    \\ \cline{4-10} 
                       &                            &                      & HIRE             & 56.44 $^{ 0.30\rule{0pt}{0.2cm}}$ & 58.22 $^{ 0.29}$ & 57.99 $^{ 1.30}$          & \textbf{62.56 $^{ 0.29}$} & 58.94 $^{ 0.43}$    & 58.93 $^{ 0.96}$    \\ \cline{4-10} 
                       &                            &                      & Improve          & 1.0\%        & 2.7\%        & \textbf{8.9\%}        & 3.0\%                 & 3.1\%           & 4.4\%           \\ \cline{3-10} 
                       &                            & \multirow{3}{*}{0.4} & Teacher          & 58.00 $^{ 0.38\rule{0pt}{0.2cm}}$ & 59.79 $^{ 0.55}$ & 58.24 $^{ 0.63}$          & \textbf{65.30 $^{ 0.40}$} & 60.29 $^{ 0.17}$    & 59.54 $^{ 0.92}$    \\ \cline{4-10} 
                       &                            &                      & HIRE             & 58.74 $^{ 0.65\rule{0pt}{0.2cm}}$ & 61.01 $^{ 0.29}$ & 62.97 $^{ 0.61}$          & \textbf{66.02 $^{ 0.19}$} & 63.13 $^{ 0.60}$    & 62.21 $^{ 0.90}$    \\ \cline{4-10} 
                       &                            &                      & Improve          & 1.3\%        & 2.0\%        & \textbf{8.1\%}        & 1.1\%                 & 4.7\%           & 4.5\%           \\ \cline{3-10} 
                       &                            & \multirow{3}{*}{0.6} & Teacher          & 59.97 $^{ 0.39\rule{0pt}{0.2cm}}$ & 60.89 $^{ 0.65}$ & 60.75 $^{ 0.50}$          & \textbf{66.39 $^{ 0.95}$} & 59.94 $^{ 0.31}$    & 59.95 $^{ 0.66}$    \\ \cline{4-10} 
                       &                            &                      & HIRE             & 60.53 $^{ 0.45\rule{0pt}{0.2cm}}$ & 61.79 $^{ 0.53}$ & 65.48 $^{ 0.71}$          & \textbf{66.45 $^{ 0.60}$} & 64.54 $^{ 2.06}$    & 63.66 $^{ 0.97}$    \\ \cline{4-10} 
                       &                            &                      & Improve          & 1.0\%        & 1.5\%        & \textbf{7.8\%}        & 0.1\%                 & 7.7\%           & 6.2\%           \\ \cline{3-10} 
                       &                            & \multirow{3}{*}{0.8} & Teacher          & 63.77 $^{ 0.90\rule{0pt}{0.2cm}}$ & 62.67 $^{ 0.42}$ & 66.73 $^{ 0.61}$          & \textbf{67.54 $^{ 0.78}$} & 61.10 $^{ 0.84}$    & 62.10 $^{ 0.99}$    \\ \cline{4-10} 
                       &                            &                      & HIRE             & 64.49 $^{ 0.38\rule{0pt}{0.2cm}}$ & 65.01 $^{ 0.44}$ & \textbf{71.55 $^{ 1.06}$} & 68.50 $^{ 0.78}$          & 67.68 $^{ 0.99 }$   & 67.30 $^{ 1.41}$    \\ \cline{4-10} 
                       &                            &                      & Improve          & 1.1\%        & 3.7\%        & 7.2\%                 & 1.4\%                 & \textbf{10.8\%} & 8.4\%           \\ \cline{2-10} 
                       & \multirow{12}{*}{Macro-F1} & \multirow{3}{*}{0.2} & Teacher          & 55.63 $^{ 0.38\rule{0pt}{0.2cm}}$ & 56.63 $^{ 0.31}$ & 53.46 $^{ 0.54}$          & \textbf{60.73 $^{ 0.31}$} & 57.16 $^{ 0.23}$    & 56.44 $^{ 0.43 }$   \\ \cline{4-10} 
                       &                            &                      & HIRE             & 56.23 $^{ 0.29\rule{0pt}{0.2cm}}$ & 57.95 $^{ 0.35}$ & 58.01 $^{ 1.30}$          & \textbf{62.54 $^{ 0.33}$} & 59.07 $^{ 0.46}$    & 59.01 $^{ 0.99}$    \\ \cline{4-10} 
                       &                            &                      & Improve          & 1.1\%        & 2.3\%        & \textbf{8.5\%}        & 3.0\%                 & 3.3\%           & 4.6\%           \\ \cline{3-10} 
                       &                            & \multirow{3}{*}{0.4} & Teacher          & 57.75 $^{ 0.38\rule{0pt}{0.2cm}}$ & 59.54 $^{ 0.48}$ & 58.24 $^{ 0.58}$          & \textbf{65.03 $^{ 0.33}$} & 60.26 $^{ 0.16}$    & 59.41 $^{ 0.86}$    \\ \cline{4-10} 
                       &                            &                      & HIRE             & 58.51 $^{ 0.63\rule{0pt}{0.2cm}}$ & 60.54 $^{ 0.25}$ & 62.84 $^{ 0.61}$          & \textbf{65.63 $^{ 0.12}$} & 63.11 $^{ 0.60}$    & 62.11 $^{ 0.86}$    \\ \cline{4-10} 
                       &                            &                      & Improve          & 1.3\%        & 1.7\%        & \textbf{7.9\%}        & 0.9\%                 & 4.7\%           & 4.6\%           \\ \cline{3-10} 
                       &                            & \multirow{3}{*}{0.6} & Teacher          & 59.81 $^{ 0.36\rule{0pt}{0.2cm}}$ & 60.75 $^{ 0.65}$ & 60.85 $^{ 0.45}$          & \textbf{66.25 $^{ 0.96}$} & 60.08 $^{ 0.32}$    & 60.10 $^{ 0.64}$    \\ \cline{4-10} 
                       &                            &                      & HIRE             & 60.32 $^{ 0.45\rule{0pt}{0.2cm}}$ & 61.61 $^{ 0.51}$ & 65.43 $^{ 0.99}$          & \textbf{66.23 $^{ 0.56}$} & 64.64 $^{ 2.02}$    & 63.75 $^{ 0.94}$    \\ \cline{4-10} 
                       &                            &                      & Improve          & 0.9\%        & 1.4\%        & 7.5\%                 & 0.0\%                 & \textbf{7.6\%}  & 6.1\%           \\ \cline{3-10} 
                       &                            & \multirow{3}{*}{0.8} & Teacher          & 63.45 $^{ 0.93\rule{0pt}{0.2cm}}$ & 62.35 $^{ 0.37}$ & 66.70 $^{ 0.58}$          & \textbf{67.35 $^{ 0.79}$} & 61.01 $^{ 0.82}$    & 62.07 $^{ 0.95}$    \\ \cline{4-10} 
                       &                            &                      & HIRE             & 64.09 $^{ 0.35\rule{0pt}{0.2cm}}$ & 64.57 $^{ 0.50}$ & \textbf{71.46 $^{ 1.06}$} & 68.33 $^{ 0.75}$          & 67.54 $^{ 1.00}$    & 67.16 $^{ 1.41}$   \\ \cline{4-10} 
                       &                            &                      & Improve          & 1.0\%        & 3.6\%        & 7.1\%                 & 1.5\%                 & \textbf{10.7\%} & 8.2\%           \\ \hline
\end{tabular}

}

\label{tab:clf-imdb}
\end{table*}

\begin{table*}[htbp]
	\caption{HIRE improves various teacher models with satisfactory margins on DBLP}. 
\centering
\resizebox{0.9\textwidth}{!}{
	
\begin{tabular}{|c|c|c|c|c|c|c|c|c|c|}
\hline
\textbf{Datasets}      & \textbf{Metrics}           & \textbf{Training}    & \textbf{Network} & \textbf{GCN} & \textbf{GAT} & \textbf{RGCN}  & \textbf{HAN}          & \textbf{HGT} & \textbf{HGConv}       \\ \hline
\multirow{24}{*}{DBLP} & \multirow{12}{*}{Micro-F1} & \multirow{3}{*}{0.2} & Teacher          & 91.81 $^{ 0.17\rule{0pt}{0.2cm}}$ & 90.85 $^{ 0.96}$ & 84.47 $^{ 1.60}$   & 92.10 $^{ 0.36}$          & 93.08 $^{ 0.30}$ & \textbf{93.38 $^{ 0.34}$} \\ \cline{4-10} 
                       &                            &                      & HIRE             & 92.05 $^{ 0.11\rule{0pt}{0.2cm}}$ & 91.73 $^{ 0.37}$ & 92.56 $^{ 0.53}$   & 92.61 $^{ 0.35}$          & 93.57 $^{ 0.40}$ & \textbf{94.50 $^{ 0.18}$} \\ \cline{4-10} 
                       &                            &                      & Improve          & 0.3\%        & 1.0\%        & \textbf{9.6\%} & 0.6\%                 & 0.5\%        & 1.2\%                 \\ \cline{3-10} 
                       &                            & \multirow{3}{*}{0.4} & Teacher          & 91.47 $^{ 0.16\rule{0pt}{0.2cm}}$ & 90.52 $^{ 0.76}$ & 88.96 $^{ 0.86}$   & 92.46 $^{ 0.44}$          & 93.17 $^{ 0.25}$ & \textbf{93.21 $^{ 0.28}$} \\ \cline{4-10} 
                       &                            &                      & HIRE             & 91.80 $^{ 0.05\rule{0pt}{0.2cm}}$ & 91.47 $^{ 0.47}$ & 93.57 $^{ 0.20}$   & 92.81 $^{ 0.21}$          & 94.29 $^{ 0.24}$ & \textbf{94.36 $^{ 0.09}$} \\ \cline{4-10} 
                       &                            &                      & Improve          & 0.4\%        & 1.0\%        & \textbf{5.2\%} & 0.4\%                 & 1.2\%        & 1.2\%                 \\ \cline{3-10} 
                       &                            & \multirow{3}{*}{0.6} & Teacher          & 92.70 $^{ 0.23\rule{0pt}{0.2cm}}$ & 91.50 $^{ 0.21}$ & 91.41 $^{ 0.61}$   & 92.97 $^{ 0.48}$          & 93.21 $^{ 0.27}$ & \textbf{93.39 $^{ 0.16}$} \\ \cline{4-10} 
                       &                            &                      & HIRE             & 93.00 $^{ 0.21\rule{0pt}{0.2cm}}$ & 92.20 $^{ 0.49}$ & 94.11 $^{ 0.26}$   & 93.57 $^{ 0.46}$          & 94.83 $^{ 0.34}$ & \textbf{95.58 $^{ 0.32}$} \\ \cline{4-10} 
                       &                            &                      & Improve          & 0.3\%        & 0.8\%        & \textbf{3.0\%} & 0.6\%                 & 1.7\%        & 2.3\%                 \\ \cline{3-10} 
                       &                            & \multirow{3}{*}{0.8} & Teacher          & 92.52 $^{ 0.38\rule{0pt}{0.2cm}}$ & 91.55 $^{ 0.51}$ & 90.63 $^{ 0.32}$   & 94.00 $^{ 0.64}$          & 93.96 $^{ 0.74}$ & \textbf{94.03 $^{ 0.00}$} \\ \cline{4-10} 
                       &                            &                      & HIRE             & 92.60 $^{ 0.38\rule{0pt}{0.2cm}}$ & 92.04 $^{ 0.45}$ & 94.22 $^{ 0.41}$   & 94.31 $^{ 0.31}$          & 95.14 $^{ 0.21}$ & \textbf{95.40 $^{ 0.14}$} \\ \cline{4-10} 
                       &                            &                      & Improve          & 0.1\%        & 0.5\%        & \textbf{4.0\%} & 0.3\%                 & 1.3\%        & 1.5\%                 \\ \cline{2-10} 
                       & \multirow{12}{*}{Macro-F1} & \multirow{3}{*}{0.2} & Teacher          & 91.20 $^{ 0.16\rule{0pt}{0.2cm}}$ & 90.31 $^{ 0.96}$ & 83.90 $^{ 1.61}$   & 91.56 $^{ 0.35}$          & 92.42 $^{ 0.39}$ & \textbf{92.82 $^{ 0.35}$} \\ \cline{4-10} 
                       &                            &                      & HIRE             & 91.47 $^{ 0.11\rule{0pt}{0.2cm}}$ & 91.16 $^{ 0.36}$ & 92.01 $^{ 0.54}$   & 92.10 $^{ 0.41}$          & 92.98 $^{ 0.13}$ & \textbf{94.05 $^{ 0.17}$} \\ \cline{4-10} 
                       &                            &                      & Improve          & 0.3\%        & 0.9\%        & \textbf{9.7\%} & 0.6\%                 & 0.6\%        & 1.3\%                 \\ \cline{3-10} 
                       &                            & \multirow{3}{*}{0.4} & Teacher          & 90.80 $^{ 0.15\rule{0pt}{0.2cm}}$ & 89.89 $^{ 0.79}$ & 88.38 $^{ 0.85}$   & 91.89 $^{ 0.46}$          & 92.53 $^{ 0.35}$ & \textbf{92.63 $^{ 0.32}$} \\ \cline{4-10} 
                       &                            &                      & HIRE             & 91.20 $^{ 0.08\rule{0pt}{0.2cm}}$ & 90.86 $^{ 0.46}$ & 93.03 $^{ 0.22}$   & 92.29 $^{ 0.21}$          & 93.77 $^{ 0.27}$ & \textbf{93.86 $^{ 0.09}$} \\ \cline{4-10} 
                       &                            &                      & Improve          & 0.4\%        & 1.1\%        & \textbf{5.3\%} & 0.4\%                 & 1.3\%        & 1.3\%                 \\ \cline{3-10} 
                       &                            & \multirow{3}{*}{0.6} & Teacher          & 91.98 $^{ 0.25\rule{0pt}{0.2cm}}$ & 90.91 $^{ 0.24}$ & 90.57 $^{ 0.60}$   & 92.41 $^{ 0.50}$          & 92.41 $^{ 0.34}$ & \textbf{92.93 $^{ 0.18}$} \\ \cline{4-10} 
                       &                            &                      & HIRE             & 92.34 $^{ 0.27\rule{0pt}{0.2cm}}$ & 91.57 $^{ 0.52}$ & 93.49 $^{ 0.25}$   & 93.05 $^{ 0.50}$          & 94.23 $^{ 0.35}$ & \textbf{95.09 $^{ 0.37}$} \\ \cline{4-10} 
                       &                            &                      & Improve          & 0.4\%        & 0.7\%        & \textbf{3.2\%} & 0.7\%                 & 2.0\%        & 2.3\%                 \\ \cline{3-10} 
                       &                            & \multirow{3}{*}{0.8} & Teacher          & 92.08 $^{ 0.42\rule{0pt}{0.2cm}}$ & 91.23 $^{ 0.55}$ & 89.87 $^{ 0.32}$   & \textbf{93.72 $^{ 0.72}$} & 93.45 $^{ 0.87}$ & 93.36 $^{ 0.03}$          \\ \cline{4-10} 
                       &                            &                      & HIRE             & 92.18 $^{ 0.40\rule{0pt}{0.2cm}}$ & 91.63 $^{ 0.49}$ & 93.88 $^{ 0.46}$   & 94.06 $^{ 0.35}$          & 94.81 $^{ 0.26}$ & \textbf{95.20 $^{ 0.16}$} \\ \cline{4-10} 
                       &                            &                      & Improve          & 0.1\%        & 0.4\%        & \textbf{4.5\%} & 0.4\%                 & 1.5\%        & 2.0\%                 \\ \hline
\end{tabular}

}

\label{tab:clf-dblp}
\end{table*}

\textbf{Results and Analysis.}
HIRE, performed on both homogeneous and heterogeneous models, has achieved considerable improvements under different training data ratios, significantly increasing $0.1\%\sim 10.8\%$ over their corresponding teacher models on the three datasets, indicating the effectiveness of our method. Additional exciting observations can be summarized as follows: 1) HIRE  varies slightly on different datasets. For example, our HIRE shows the most significant improvement under the 20\% training ratio in ACM, proving the effectiveness of HIRE in data scarcity scenarios. What is more, in IMDB and DBLP, HIRE can continuously improve their teacher models as the training data increases and get the most remarkable relative improvement when the data ratio is 80\%. 2) RGCN benefits most from our HIRE framework. For example, in ACM, RGCN can achieve even competitive results or better than HGT. One possible reason is that it is difficult for RGCN to deal with the noise and complex patterns in the raw heterogeneous graphs. In contrast, the supervised information refined by HIRE can effectively guide the training of RGCN. Similarly, in IMDB and DBLP, RGCN still benefits the most from HIRE, increasing $7.1\%\sim8.9\%$ and $3.0\%\sim9.6\%$ respectively. 3) Under the guidance of HIRE, the improvement of GCN, GAT, and HAN is not as good as the other three heterogeneous models. For example, in ACM, their improvement ranges are $0.1\%\sim3.7\%$ and $3.1\%\sim10.8\%$, respectively. GCN, GAT, and HAN only use NKD, especially the model capacity of GCN and GAT are not enough for node representation on heterogeneous graphs. 4) Meanwhile, HIRE can continuously improve the baseline of the strongest teacher model as the ratio of training data increases. For example, guided by HIRE, HGConv enhances the performance from 93.38\% to 94.50\% when the training data ratio is 20\% in DBLP. HAN has a similar performance in IMDB, whose performance is increased from 60.74\% to 62.56\% at a 20\%training data ratio.

\subsubsection{Node Clustering}
In the control experiments on node clustering tasks, the comparative analysis with and without HIRE is also carried out on ACM, IMDB, and DBLP datasets. The overall results are summarized in Table~\ref{tab:clu}, which are reported as the mean and variance of clustering metrics with five runs. It is obvious that the student model guided by HIRE can continuously and significantly improve the performance of their corresponding teacher model with an improved range of $0.3\%\sim53.3\%$.

\begin{table*}[htbp]
\caption{HIRE improves various teacher models with satisfactory margins on all datasets. We report the NMI and ARI on ACM, IMDB, and DBLP for the node clustering task. }
\centering
\resizebox{0.9\textwidth}{!}{
	
\begin{tabular}{|c|c|c|c|c|c|c|c|c|}
\hline
\textbf{Datasets}     & \textbf{Metrics}     & \textbf{Network} & \textbf{GCN} & \textbf{GAT} & \textbf{RGCN}   & \textbf{HAN}          & \textbf{HGT}          & \textbf{HGConv}       \\ \hline
\multirow{6}{*}{ACM}  & \multirow{3}{*}{NMI} & Teacher          & 61.73 $^{ 1.37\rule{0pt}{0.2cm}}$ & 64.63 $^{ 1.23}$ & 66.16 $^{ 2.98}$    & 68.08 $^{ 2.25}$          & \textbf{69.69 $^{ 1.38}$} & 67.33 $^{ 3.26}$          \\ \cline{3-9} 
                      &                      & HIRE             & 62.69 $^{ 0.62\rule{0pt}{0.2cm}}$ & 65.51 $^{ 0.89}$ & 71.54 $^{ 2.17}$    & 71.51 $^{ 0.84}$          & 70.88 $^{ 1.07}$          & \textbf{73.12 $^{ 1.26}$} \\ \cline{3-9} 
                      &                      & Improve          & 1.6\%           & 1.4\%           & 8.1\%           & 5.0\%                 & 1.7\%                 & \textbf{8.6\%}        \\ \cline{2-9} 
                      & \multirow{3}{*}{ARI} & Teacher          & 65.73 $^{ 1.51\rule{0pt}{0.2cm}}$ & 69.86 $^{ 0.87}$ & 70.24 $^{ 2.28}$    & 73.69 $^{ 1.83}$          & \textbf{74.96 $^{ 1.54}$} & 73.22 $^{ 2.96} $         \\ \cline{3-9} 
                      &                      & HIRE             & 67.24 $^{ 0.53\rule{0pt}{0.2cm}}$ & 70.04 $^{ 0.71}$ & 75.43 $^{ 2.41}$    & 76.53 $^{ 0.89}$          & 76.20 $^{ 0.97}$          & \textbf{78.14 $^{ 1.09}$} \\ \cline{3-9} 
                      &                      & Improve          & 2.3\%           & 0.3\%           & \textbf{7.4\%}  & 3.9\%                 & 1.7\%                 & 6.7\%                 \\ \hline
\multirow{6}{*}{IMDB} & \multirow{3}{*}{NMI} & Teacher          & 10.06 $^{ 0.39\rule{0pt}{0.2cm}}$ & 11.31 $^{ 0.32}$ & 8.94 $^{ 0.32}$     & \textbf{14.91 $^{ 0.24}$} & 11.90 $^{ 0.19}$          & 10.81 $^{ 0.43}$          \\ \cline{3-9} 
                      &                      & HIRE             & 10.40 $^{ 0.17\rule{0pt}{0.2cm}}$ & 12.46 $^{ 0.10}$ & 12.20 $^{ 0.96}$     & \textbf{16.05 $^{ 0.26}$} & 14.10 $^{ 0.56}$          & 12.90 $^{ 0.93}$          \\ \cline{3-9} 
                      &                      & Improve          & 3.4\%           & 10.2\%          & \textbf{36.5\%} & 7.6\%                 & 18.5\%                & 19.3\%                \\ \cline{2-9} 
                      & \multirow{3}{*}{ARI} & Teacher          & 11.35 $^{ 0.64\rule{0pt}{0.2cm}}$ & 12.27 $^{ 0.38}$ & 9.52 $^{ 0.31}$     & \textbf{16.24 $^{ 0.32}$} & 13.36 $^{ 0.20}$          & 12.08 $^{ 0.37}$          \\ \cline{3-9} 
                      &                      & HIRE             & 11.84 $^{ 0.19\rule{0pt}{0.2cm}}$ & 13.15 $^{ 0.42}$ & 12.69 $^{ 0.64}$    & \textbf{17.29 $^{ 0.33}$} & 15.43 $^{ 0.79}$          & 13.58 $^{ 2.8 } $         \\ \cline{3-9} 
                      &                      & Improve          & 4.3\%           & 7.2\%           & \textbf{33.3\%} & 6.5\%                 & 15.5\%                & 12.4\%                \\ \hline
\multirow{6}{*}{DBLP} & \multirow{3}{*}{NMI} & Teacher          & 70.88 $^{ 0.85\rule{0pt}{0.2cm}}$ & 71.16 $^{ 1.25}$ & 52.32 $^{ 7.04}$    & 75.75 $^{ 0.31}$          & 77.56 $^{ 0.74}$          & \textbf{78.47 $^{ 0.37}$} \\ \cline{3-9} 
                      &                      & HIRE             & 70.33 $^{ 0.43\rule{0pt}{0.2cm}}$ & 70.99 $^{ 2.15}$ & 78.78 $^{ 0.25}$    & 76.40 $^{ 0.37}$          & 78.98 $^{ 0.80}$          & \textbf{80.65 $^{ 0.35}$} \\ \cline{3-9} 
                      &                      & Improve          & -0.8\%          & -0.2\%          & \textbf{50.6\%} & 0.9\%                 & 1.8\%                 & 2.8\%                 \\ \cline{2-9} 
                      & \multirow{3}{*}{ARI} & Teacher          & 76.50 $^{ 0.71\rule{0pt}{0.2cm}}$ & 75.75 $^{ 1.73}$ & 54.96 $^{ 11.36}$   & 81.69 $^{ 0.49}$          & 83.49 $^{ 0.70}$          & \textbf{84.09 $^{ 0.41}$} \\ \cline{3-9} 
                      &                      & HIRE             & 75.94 $^{ 0.36\rule{0pt}{0.2cm}}$ & 75.37 $^{ 2.86}$ & 84.28 $^{ 0.19}$    & 82.21 $^{ 0.26}$          & 84.85 $^{ 0.62}$          & \textbf{86.18 $^{ 0.35}$} \\ \cline{3-9} 
                      &                      & Improve          & -0.7\%       & -0.5\%       & \textbf{53.3\%} & 0.6\%                 & 1.6\%                 & 2.5\%                 \\ \hline
\end{tabular}

}
\label{tab:clu}
\end{table*}

\textbf{Results and Analysis.}
Firstly, on the ACM dataset, all models based on HIRE can be improved by about $0.3\%\sim8.6\%$ of the clustering effect. Among them, HGConv achieves the best performance, followed by RGCN. Secondly, we find some interesting observations on the IMDB dataset: 1) Surprisingly, HIRE has the most significant improvement with $3.4\%\sim36.5\%$ on IMDB compared to ACM and DBLP. For example, the performance of HGConv on IMDB increases from 10.81\% to 12.90\% with the clustering metric NMI, which verifies that HIRE can obtain valuable information from extremely complex heterogeneous graphs (movie labels in IMDB datasets are not precise, as mentioned earlier). 2) HAN performs best in all teacher models, and HIRE still improves HAN by approximately 7.6\% with the NMI metric. Thirdly, on the DBLP dataset, the relative improvement range of the four heterogeneous models is $0.9\%\sim53.3\%$. Among them, HIRE increases the most robust baseline of HGConv by approximately 2.8\%, which verifies the power of our distillation method on various HGNNs. However, the two homogeneous models (i.e., GCN and GAT) show slight degradation, which may be that these models are not enough to learn complex patterns on heterogeneous graphs. Finally, models achieving significant improvements results on node classification tasks (e.g., RGCN and HGConv) also have satisfactory performance on node clustering tasks. Thus, guided by HIRE, HGNNs can learn more general node embedding that could be applied to various downstream tasks.

\subsubsection{Node Visualization}
\begin{figure*}[htbp]
	\centering
	\includegraphics[width=\textwidth]{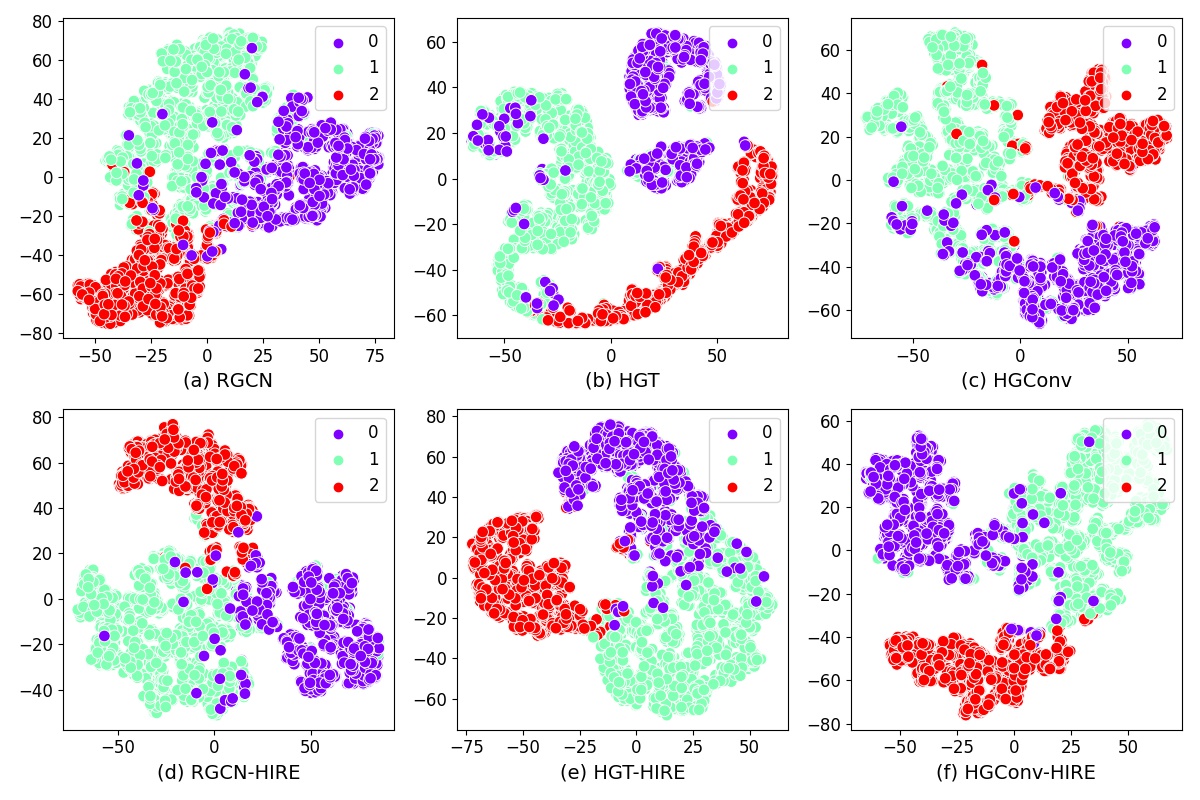}
	\caption{Visualization of node representation on ACM. Each point indicates a paper and its color denotes the published area.}
	\label{fig:nv_acm}
\end{figure*}

For qualitative comparison, we use t-SNE on ACM and DBLP\footnote{Please refer to~\ref{sec:appendixB} for results on DBLP.
} datasets (see Figure~\ref{fig:nv_acm} and Figure~\ref{fig:nv_dblp}) to visualize the target nodes paper/author in the heterogeneous graph into a 2-dimensional plane and color the nodes according to their research areas.

\textbf{Results and Analysis.}
From Figure~\ref{fig:nv_acm}, we observe that the teacher models cannot achieve satisfactory performance. They are either unable to gather nodes in the same area or provide clear boundaries for nodes belonging to different research areas, showing unsatisfactory discrimination performance. Under the guidance of HIRE, the node embedding capabilities of all models have been further significantly improved, enabling papers in the same area with higher intra-class similarity and distinguishing papers in different research areas with clear boundaries.

\subsection{RQ2: Ablation studies} 
\subsubsection{Different-order Knowledge Distillation}
\begin{figure*}[htbp]
	\centering
	\includegraphics[width=\textwidth]{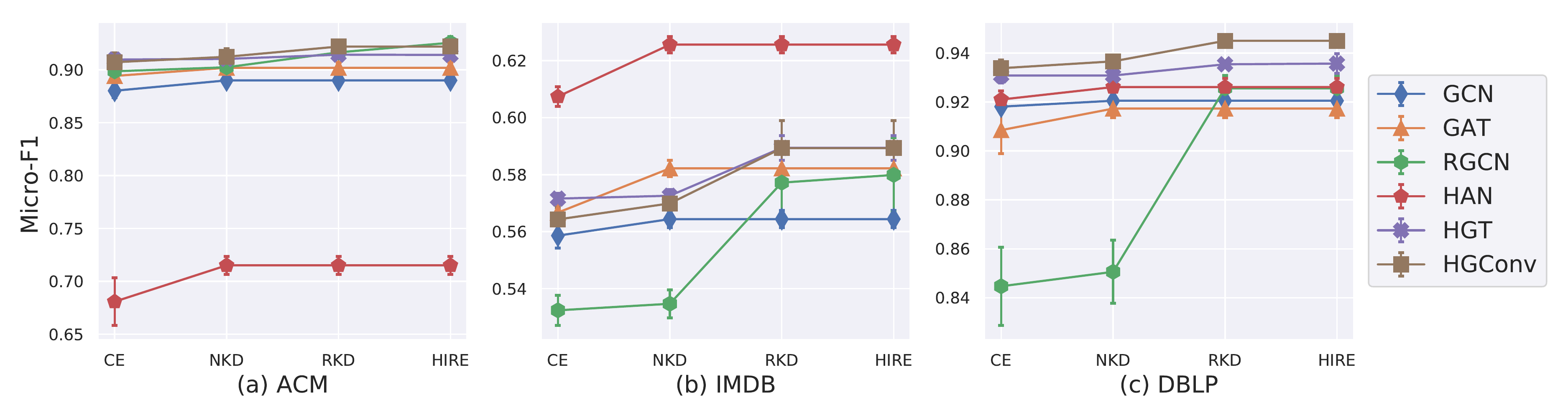}
	\caption{Effects of the components in HIRE on node classification task with 20\% training ratios. The x-axis is the variants of HIRE and the y-axis is the metric Micro-F1. }
	\label{fig:clf_variant_mi20}
\end{figure*}

\begin{figure*}[htbp]
	\centering
	\includegraphics[width=\textwidth]{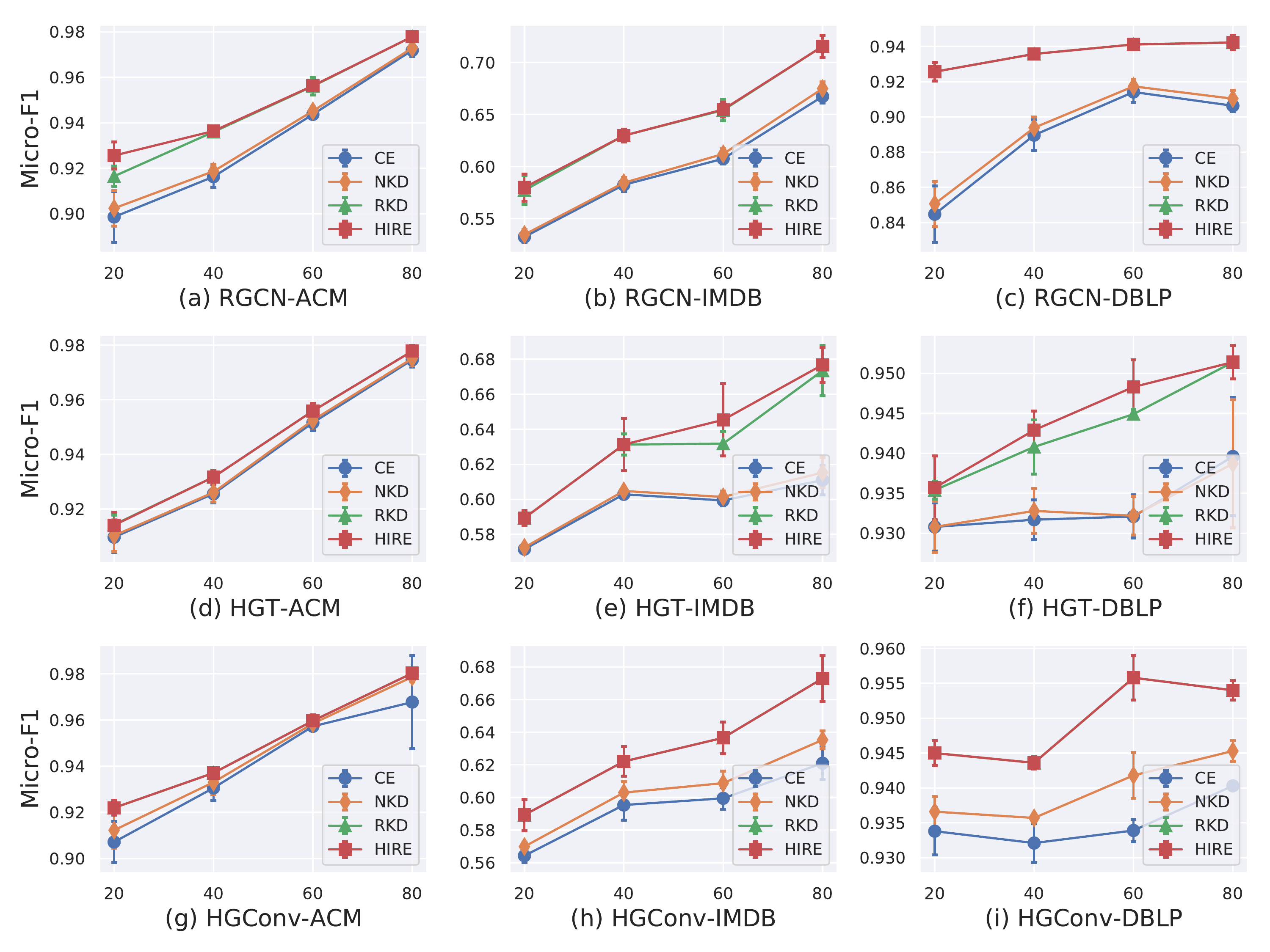}
	\caption{Effects of the components in HIRE on node classification task with different training ratios. The x-axis is the training ratio and the y-axis is the metric Micro-F1. 
	}
	\label{fig:clf_variants_mi}
\end{figure*}

To validate the effectiveness of each component of HIRE, we further conduct experiments on different HIRE variants. Experimental results of the variants on node classification task are illustrated in Figure~\ref{fig:clf_variant_mi20} and Figure~\ref{fig:clf_variants_mi}, where CE (cross-entropy) is equal to the teacher model. We add node-level knowledge distillation and relation-level knowledge distillation, respectively, and denote the two variants as NKD and RKD. Detailed implementations of the four variants are described as follows: 
\begin{itemize}
 \item \textbf{CE}. This variant adopts the ground truth labels calculated by the basic cross-entropy loss. And the overall loss in Eq.~\ref{eq:totalloss} becomes $\mathcal{L} = \mathcal{L} _{CE} $.
 
 \item \textbf{NKD}. This variant applies node-level knowledge distillation denoted in Eq.~\ref{eq:kdloss} 

 \item \textbf{RKD}. This variant replaces the node-level knowledge distillation with the relation-level knowledge distillation to imitate the relational correlation between node embeddings of different types generated by the teacher HGNN. Meanwhile, the total loss becomes $\mathcal{L} = \mathcal{L} _{CE} + \beta \ast \mathcal{L} _{RKD} $
 
 \item \textbf{HIRE}. This variant is our proposed framework using the node-level knowledge distillation and the relation-level knowledge distillation, which is denoted in Eq.~\ref{eq:totalloss}.
\end{itemize}
Except for the differences mentioned above, all other settings are the same for these HIRE variants. For GCN, GAT, and HAN, HIRE degenerates the NKD variant because they all convert heterogeneous graphs into homogeneous graphs before message passing. Therefore, there are no multi-type node representations in the hidden layer of these models. Thus the second-order relational-level knowledge distillation degenerates into the first-order node-level knowledge distillation in these models. To intuitively analyze the effect of RKD in HIRE, we conducted a variety of variant experiments in models using RKD, including RGCN, HGT, and HGConv.

\textbf{Results and Analysis.}
 As demonstrated in Figure~\ref{fig:clf_variant_mi20}, we first conduct node classification tasks for all GNN and HGNN models under a standard 20\% training ratio to explore the effects of the components in HIRE. It can be seen that the impact of NKD, RKD, and HIRE vary in different datasets, but they all contribute to the improvements in the classification performance. Overall, NKD contributes a relatively weak and limited improvement, RKD obtains a significant improvement, and HIRE outperforms NKD and RKD.
 The results demonstrate that it is essential to distill knowledge from both node-level and relation-level knowledge distillation inside HGNNs.

Next, we analyze the effects of the components in HIRE on node classification tasks under different training ratios for RGCN, HGT, and HGConv. As shown in Figure~\ref{fig:clf_variants_mi}, the performance of the four variants of HIRE basically continues to increase as the training ratio increases, and HIRE remains the best, showing $ HIRE\ge RKD\gg NKD> CE$. On one hand, compared with NKD, RKD boosts performance with significant margins, which validates its effectiveness. On the other hand, without NKD or RKD, the corresponding variants (RKD/NKD) perform worse than HIRE, indicating the importance of simultaneously modeling NKD and RKD. 
Moreover, for DBLP, CE and NKD variants of RGCN have slightly worse performance when the training ratio is 80\%, and HGConv also has a similar situation, which may be caused by overfitting. More results of HIRE's different variants can be found in~\ref{sec:appendixC}.

\begin{figure*}[htbp]
	\centering
	\includegraphics[width=\textwidth]{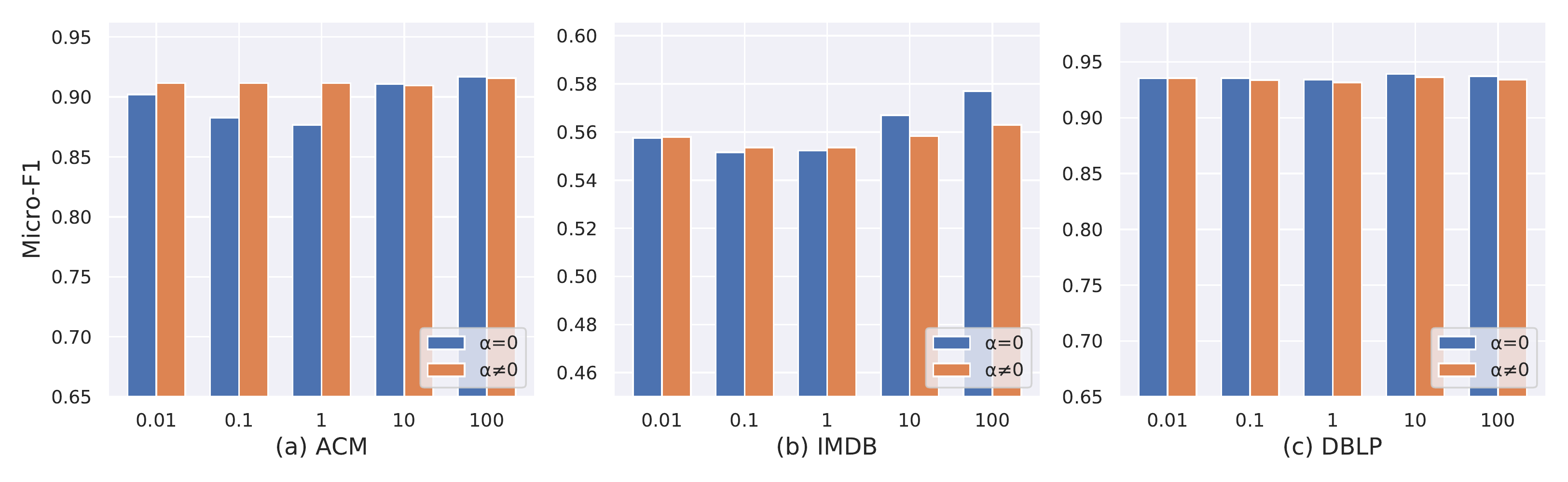}
	\caption{The impact of NKD in HIRE on node classification task. The x-axis is the values of $\beta$ and the y-axis is the metric Micro-F1.}
	\label{fig:bar-rkd-hire}
\end{figure*}

\begin{figure*}[htbp]
	\centering
	\includegraphics[width=\textwidth]{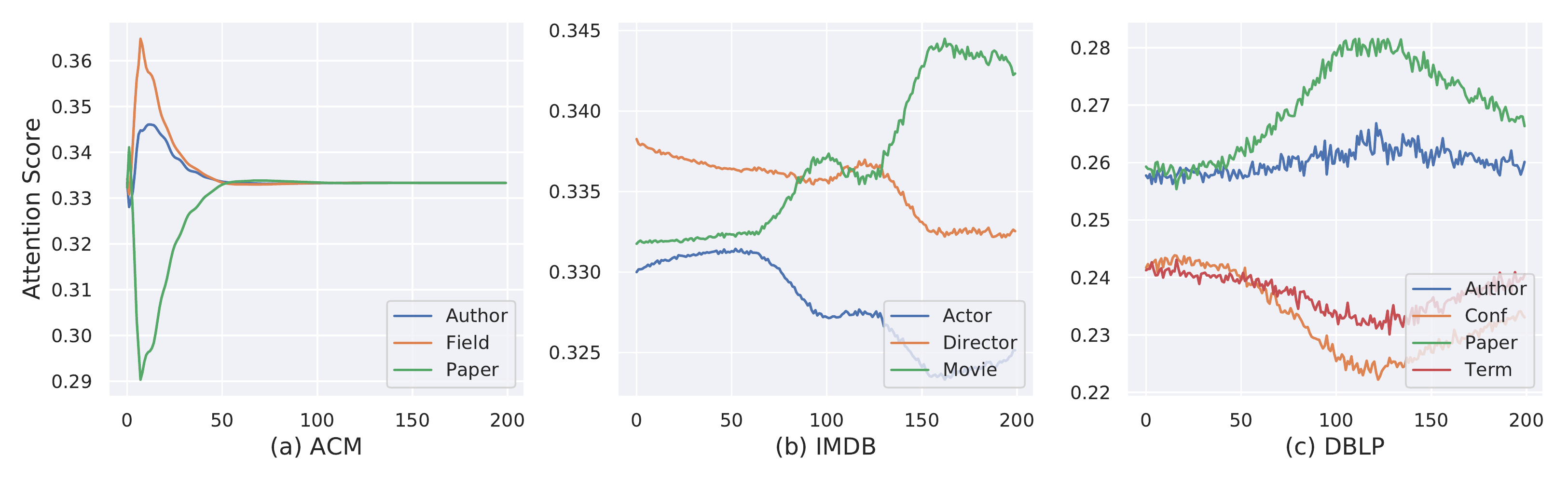}
	\caption{The attention values of node types w.r.t epochs on HGConv. }
	\label{fig.att_hgconv}
\end{figure*}
\textbf{Interesting Results and Analysis.}
Further, we explore the effect of NKD (see Figure~\ref{fig:bar-rkd-hire} ) on HIRE. Taking HGConv on ACM, IMDB, and DBLP for node classification as examples, we report the classification results (Micro-F1) of HIRE without and with NKD, which corresponds to $\alpha = 0$ and $\alpha \ne 0$, respectively. Obviously, NKD has a positive effect on HIRE in most cases. For ACM drawn in Figure~\ref{fig:bar-rkd-hire}(a), HIRE with NKD outperforms its HIRE without NKD variant when $ \beta\in \left [ 0.01,1 \right ]$, which is the same for IMDB and DBLP. Interestingly, we find that NKD harms HIRE, where HGConv guided by HIRE with NKD performs worse than itself guided by HIRE without NKD when $\beta > 1$, especially on IMDB illustrated in Figure~\ref{fig:bar-rkd-hire}(b).

\subsubsection{Attention Visualization}\label{sec:rq2_att}
A notable property of HIRE is the incorporation of a type-related attention layer module, which considers the importance of different node types in heterogeneous graphs when learning representative node embedding. We hold a reasonable assumption that some vital nodes useful for the specific downstream task tend to have larger attention values. To verify the effects of the type-related attention module, we analyze the changing trends of attention values during its training process in Figure~\ref{fig.att_hgconv}, where the x-axis is the training epoch, and the y-axis is the attention values of various nodes. Here, we provide a detailed analysis of different datasets.


\textbf{Results and Analysis.}
For example, in ACM, the attention value of Author and Field increases first and then decreases in the training process, and Paper shows the contrary trend. Finally, their attention values become stable and proximal. It presents that Field is essential for the paper classification task in the early stage of model training, but as the model training continues, the node itself plays the most critical role. In the end, these three kinds of nodes are equally crucial for the paper classification task.

In IMDB, the attention value of Actor, Director, and Movie varies little at first. Then Movie becomes more prominent, and the importances of the former two node types become smaller, indicating that Movie plays a leading role in determining the type of movie, and the neighbor node (i.e., Actor and Director) is supplementary information.

In DBLP, the attention values of Author and Paper first rise and then fall, while Conf and Term demonstrate the opposite trend. It means that Conf and Term are the critical node information to determine the author's research area, whose reason may be that the author's research area is highly related to the keywords of the conference and submitted paper.

We notice that the type-related attention module designed for RKD can reveal the differences between different nodes and weights them adequately. More results of attention values of other heterogeneous models are provided in  ~\ref{sec:appendixD}.
\subsection{RQ3: Parameter Sensitivity} \label{sec:rq3}
In this subsection, we tune several hyperparameters on ACM, IMDB, and DBLP datasets to validate the sensitivity of the HIRE method. More specifically, we use grid search to select the best hyperparameters of $\tau$, $\alpha$, and $\beta $ in Eq.~\ref{eq:totalloss} and report the results of classification (Micro-F1) with various HGNNs, as depicted in Figure~\ref{fig:psa_line}. In addition, we only conduct parameter sensitivity analysis on RGCN, HGT, and HGConv because these three models adopt both NKD and RKD. The detailed analysis is as follows:

\begin{figure*}[htbp]
	\centering
	\includegraphics[width=\textwidth]{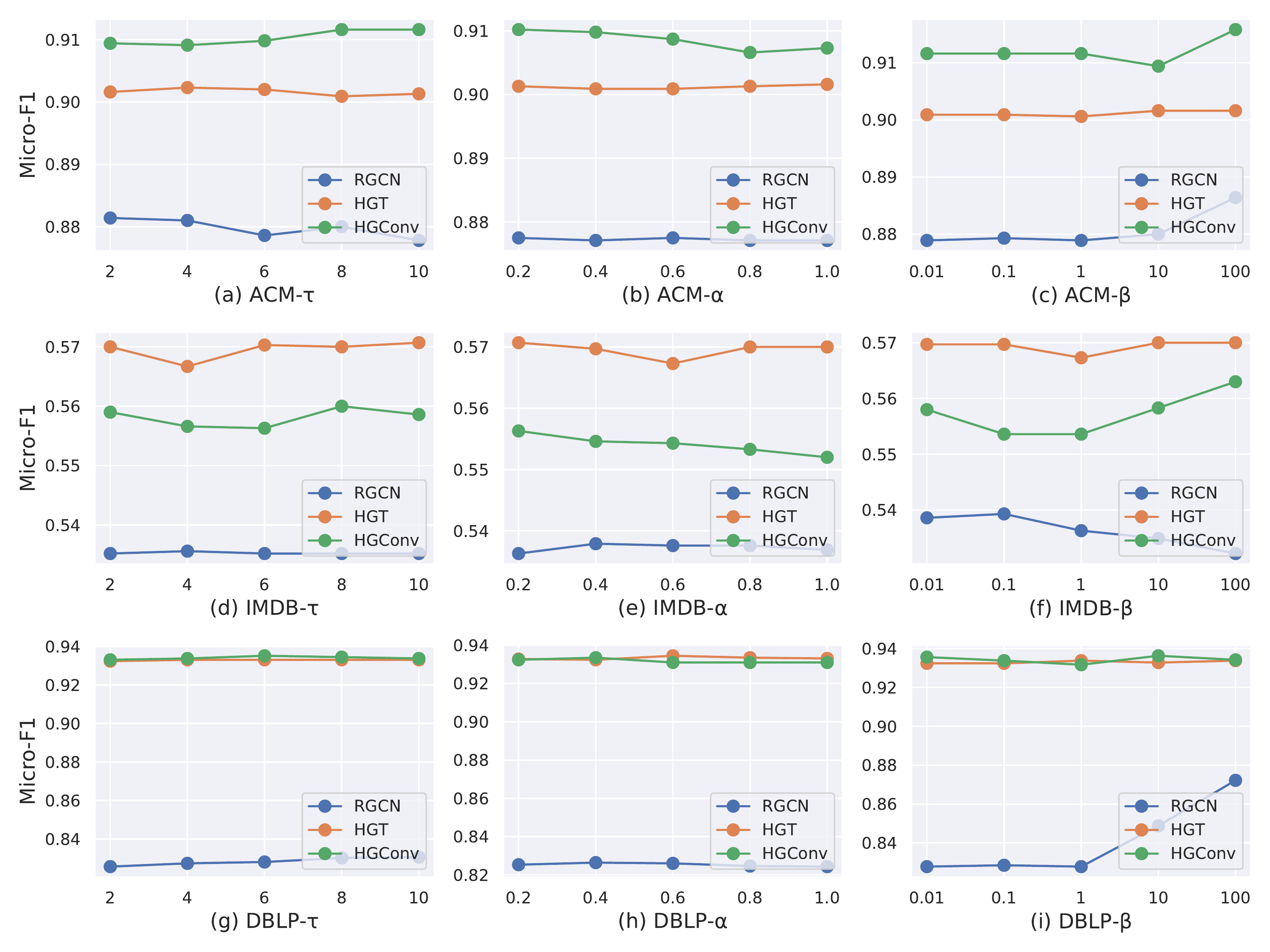}
	\caption{Parameter sensitivity of HIRE w.r.t.Number of temperature $\tau$,  Number of weighting factor $\alpha$ and  Number of weighting factor $\beta$ on ACM, IMDB, and DBLP datasets. The legends indicate the model (i.e., RGCN, HGT, and HGConv) guided by HIRE.  
	}
	\label{fig:psa_line}
\end{figure*}

\textbf{The temperature $\tau$} is adopted to scale the predicted logits and the soft labels in the distillation loss Eq.~\ref{eq:kdloss}. A higher value of $\tau$ leads to a smoother probability distribution over classes. As demonstrated in Figure~\ref{fig:psa_line}(a), (d), and (g), the performance of all models on ACM, IMDB, and DBLP datasets increases when changing $\tau$ from 2 to 8. We get the best performance in all tested models when the temperature is set to 8. When the temperature goes larger than 8, we observe a slight decrease in performance, related to the over smoothing of HGNNs.

\textbf{The weighting factor $\alpha$} is adopted to balance the cross-entropy loss and the node-level knowledge distillation term in Eq.~\ref{eq:totalloss}. We are interested in how much it affects the final performance. As illustrated in Figure~\ref{fig:psa_line}(b), (e), and (h), the primary observation is that HIRE is not sensitive to the $\alpha$ and achieves robust results on all datasets for all tested models. The results show that the HIRE method varies slightly on different models with different $\alpha$.  
For example, the performance of RGCN remains essentially unchanged across the three datasets. HGT increases with $\alpha $ on ACM, IMDB, and DBLP, obtaining the best performance when $\alpha $ is set to 1, 0.8, 0.6 respectively on the three datasets. HGConv decreases as $\alpha $ increases on the three datasets, but minor performance changes.

\textbf{The weighting factor $\beta $} is adopted to balance the node-level knowledge distillation term and the relation-level knowledge distillation term in Eq.~\ref{eq:totalloss}. To check the impact of relation-level knowledge distillation, we explore the performance of HIRE with the various values of the weighting factor $\beta $ as illustrated in Figure~\ref{fig:psa_line}(c), (f), and (i). Overall, the larger
$\beta $, the higher performance of HIRE achieves. The performance of HGT and HGConv increases as $\beta $ increases on ACM, IMDB, and DBLP, which obtains the maximum when $\beta $ is set as 100. When $ \beta\in \left [ 0.01,1 \right ]$, the performance of RGCN is stable on ACM, IMDB and DBLP datasets. However, when $ \beta\in \left [ 1,100 \right ]$, the performance of RGCN increases with the increase of $\beta $ on ACM and DBLP and is reversed in the IMDB, but the performance change of RGCN is not significant.

\section{Conclusion}\label{sec:conclusion}
In this paper, we propose the \textbf{HI}gh-order \textbf{RE}lational (\textbf{HIRE}) knowledge distillation framework based on HGNNs for the first time, filling the gap in extracting knowledge from heterogeneous graph models. HIRE combines first-order node-level knowledge and second-order relation-level knowledge, which is flexible and applied on an arbitrary HGNN. Primarily, we introduce the relation-level knowledge named RKD to retain the inherent correlation between different types of nodes in a heterogeneous graph to extract high-level knowledge hidden in HGNNs. In this way, the student model can take advantage of RKD and thus significantly outperform the teacher HGNN. Extensive experiments conducted on ACM, IMDB, and DBLP datasets prove the effectiveness of our proposed HIRE. In future work, we consider studying self-distillation on the heterogeneous graph, e.g., following~\cite{Chen2021OnSG}. Self-distillation is a promising and much more efficient training technique, aims at transferring the knowledge hidden in itself without an additional pre-trained teacher model, and our HIRE can be easily applied to self-distillation technology. Moreover, in addition to being applied to image classification~\cite{Xue2021KDExplainerAT}, relation extraction~\cite{Zhang2020DistillingKF}, and product recommendation~\cite{Wang2019BinarizedCF}, our HIRE can also be extended to GNN-based EEG applications~\cite{Kose2021FeatureEA,zhong2020eeg,song2018eeg}, while is not applicable for graph theoretical features of EEG applications~\cite{bullmore2009complex,de2009functional}, which will be explored in the future.

\textbf{CRediT authorship contribution statement}

\textbf{Jing Liu:} Investigation, Resources, Formal analysis, Methodology, Software, Validation, Visualization, Data curation, Writing - original \& draft. \textbf{Tongya Zheng:} Supervision, Conceptualization, Methodology, Software, Writing - review \& editing. \textbf{Qinfen Hao:} Writing - review.

\textbf{Declaration of Competing Interest}

The authors declare that they have no known competing financial interests or personal relationships that could have appeared to influence the work reported in this paper.

\appendix
\section{Implementation details of models}\label{sec:appendixA}
To ensure the experiment's reproducibility, we provide details of the experimental settings of the baseline models. For these six models, we follow the settings in their original paper. The details are shown in Table~\ref{tab:model}.

\begin{table}[htbp]
\caption{Experimental settings of all models used in this paper. - indicates that the attention mechanism is not used on the model.}
	\centering 
	\resizebox{1.0\linewidth}{!}{ 
    \begin{tabular}{|c|c|c|c|c|c|c|}
    \hline
    Model  & Optimizer & Learning rate & Weight decay & Dropout & Hidden layer size & Attention heads \\ \hline
    GCN    & Adam      & 0.01          & 5e-4         & 0.5     & 16                & -               \\ \hline
    GAT    & Adam      & 0.05          & 5e-4         & 0.6     & 64                & 8               \\ \hline
    RGCN   & Adam      & 0.01          & 5e-4         & 0.0     & 16                & -               \\ \hline
    HAN    & Adam      & 0.05          & 1e-3         & 0.6     & 64                & 8               \\ \hline
    HGT    & AdamW     & 0.001         & 1e-3          & 0.5     & 256               & 8               \\ \hline
    HGConv & Adam      & 0.001         & 1e-3         & 0.5     & 512               & 8               \\ \hline
    \end{tabular}
    	}
	\label{tab:model}
    
\end{table}

\section{Node visualization on DBLP}\label{sec:appendixB}
\begin{figure*}[htbp]
	\centering
	\includegraphics[width=\textwidth]{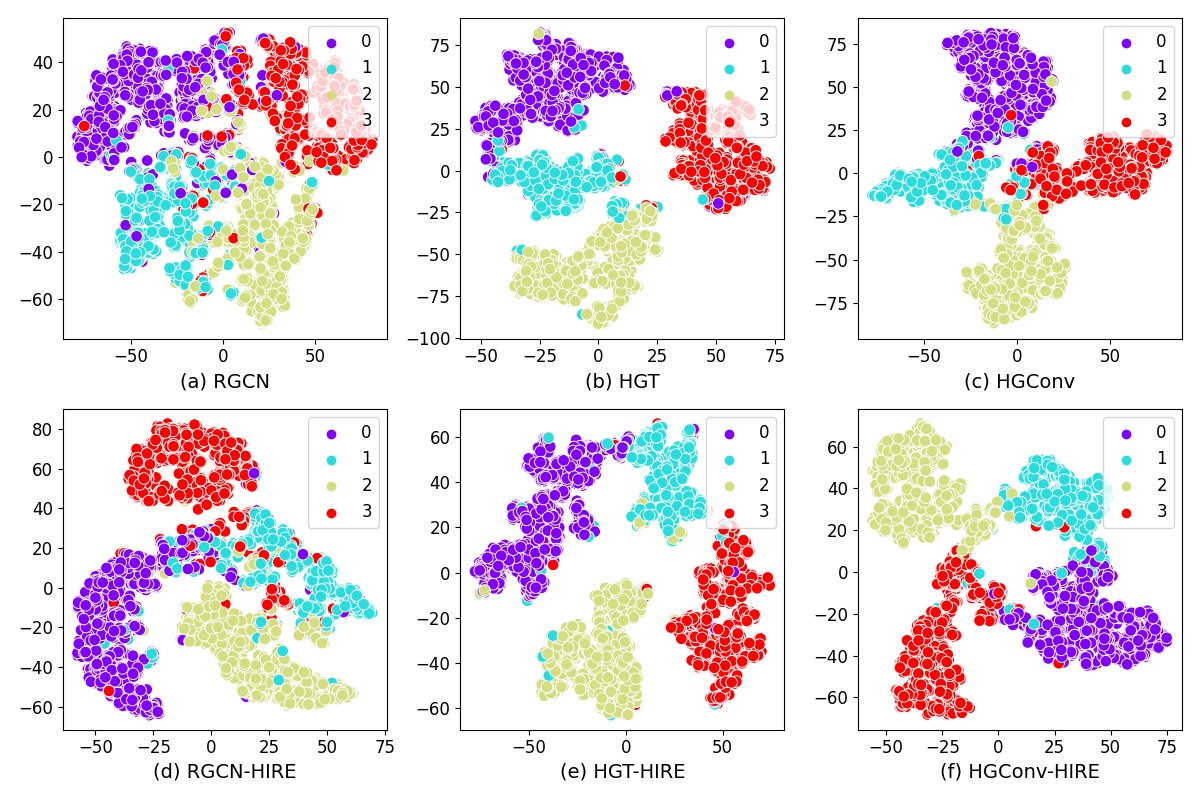}
	\caption{Visualization of node representation on DBLP. Each point indicates one author and its color indicates the research area.}
	\label{fig:nv_dblp}
\end{figure*}

For a more intuitive comparison, we also conduct the task of visualization on DBLP. We utilize t-SNE to visualize the author embedding in DBLP and color the nodes based on their research areas. From Figure~\ref{fig:nv_dblp}, we discover that RGCN does not perform well, where the authors belonging to different research areas are mixed. HGConv's boundary is blurry. HGT performs much better than RGCN and HGT. All the models guided by HIRE perform better in the node visualization, where authors within the same area are closer and boundaries between different areas are clearer.

\section{Ablation studies of Different-order Knowledge Distillation}\label{sec:appendixC}
\begin{figure*}[htbp]
	\centering
	\includegraphics[width=\textwidth]{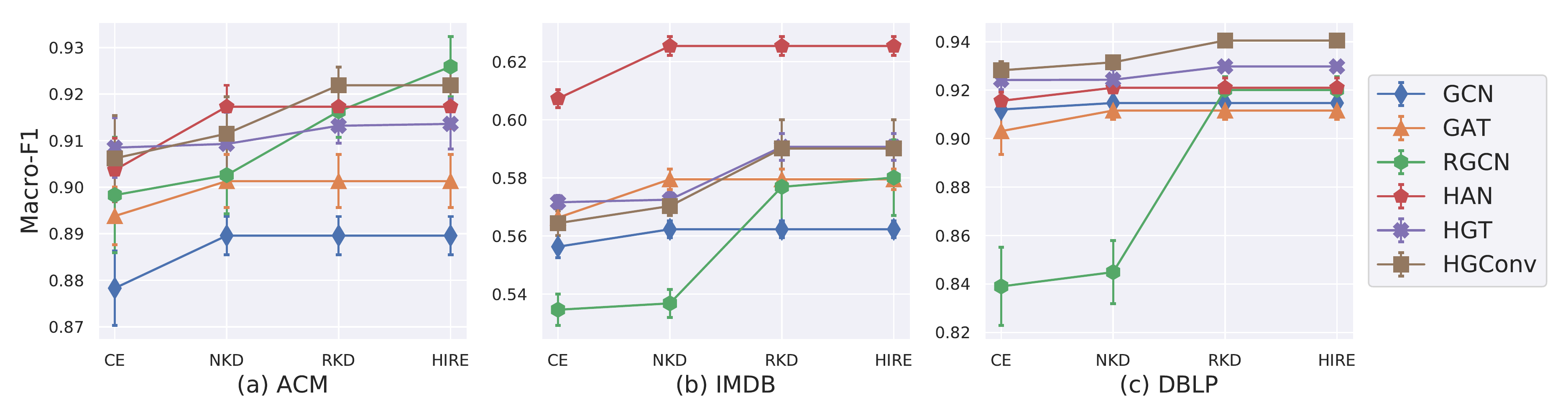}
	\caption{Effects of the components in HIRE on node classification task with 20\% training ratio. The x-axis is the variants of HIRE and the y-axis is the metric Macro-F1.}
	\label{fig:clf_variant_ma20}
\end{figure*}

\begin{figure*}[htbp]
	\centering
	\includegraphics[width=\textwidth]{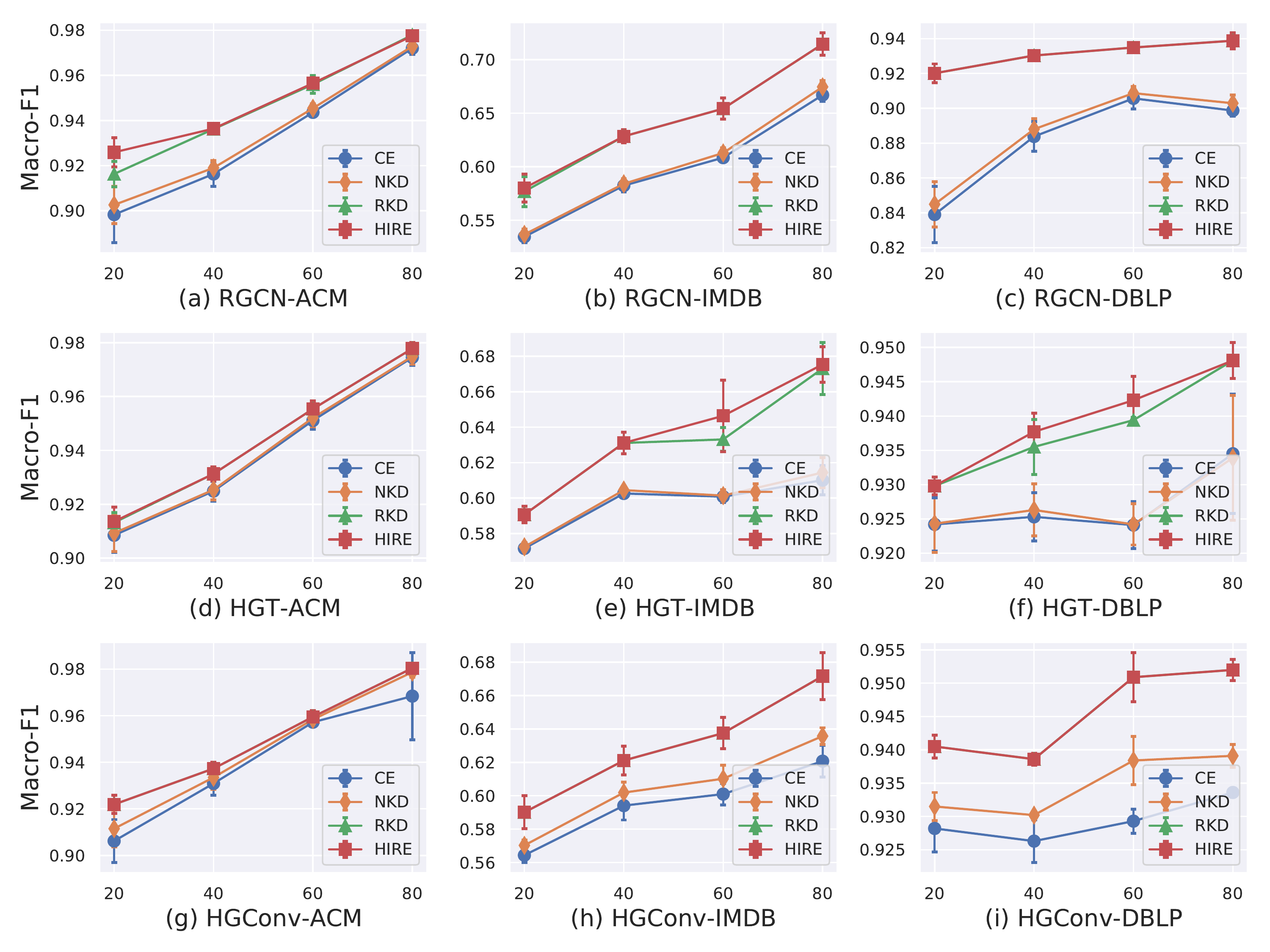}
	\caption{Effects of the components in HIRE on node classification task with different training ratios. The x-axis is the training ratio and the y-axis is the metric Macro-F1. }
	\label{fig:clf_variants_ma}
\end{figure*}

\begin{figure*}[htbp]
	\centering
	\includegraphics[width=\textwidth]{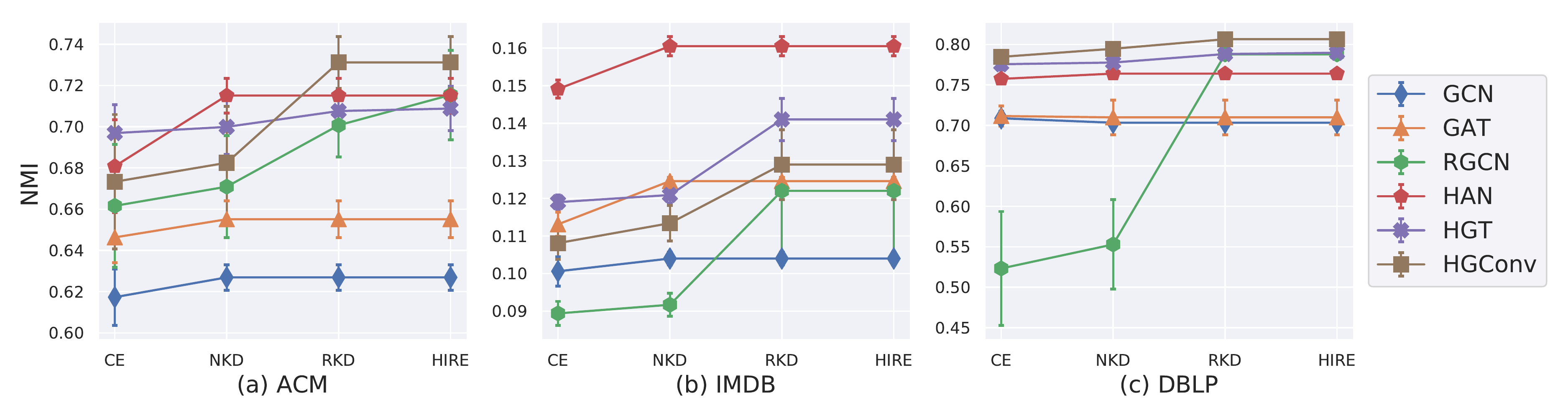}
	\caption{Effects of the components in HIRE on node clustering task with 20\% training ratio. The x-axis is the variants of HIRE and the y-axis is the metric NMI.}
	\label{fig:clu_variant_nmi20}
\end{figure*}

\begin{figure*}[htbp]
	\centering
	\includegraphics[width=\textwidth]{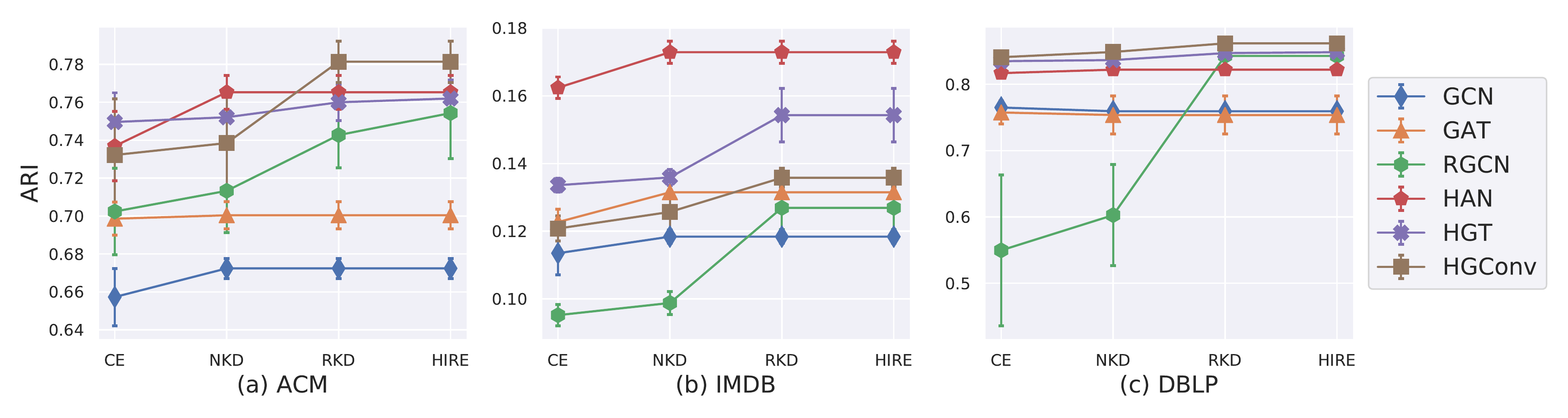}
	\caption{Effects of the components in HIRE on node clustering task with 20\% training ratio. The x-axis is the variants of HIRE and the y-axis is the metric ARI.}
	\label{fig:clu_variant_ari20}
\end{figure*}

To validate the effectiveness of each component of our model, we further conduct experiments with different HIRE variants on various models. Here we report the results obtained from the three datasets on
GNN and HGNN models in Figure~\ref{fig:clf_variant_ma20}, Figure~\ref{fig:clf_variants_ma}, Figure~\ref{fig:clu_variant_nmi20} and Figure~\ref{fig:clu_variant_ari20}. Note that every presented score of the node classification/clustering task (i.e., Macro-F1/NMI and ARI) is an average and variance of the corresponding metric (explained in Section~\ref{sec:experimentsetting}). As can be seen, RKD obtains a significant performance improvement over NKD, showing that the necessity of applying relation-level knowledge distillation to incorporate node features. Compared HIRE with CE, NKD, and RKD, HIRE beats them on all tested models, validating the efficacy of our framework. The results above demonstrate that it is quite important to distill both NKD and RKD in heterogeneous graph analysis.

\section{Attention visualization for HGT and RGCN}\label{sec:appendixD}
\begin{figure*}[htbp]
	\centering
	\includegraphics[width=\textwidth]{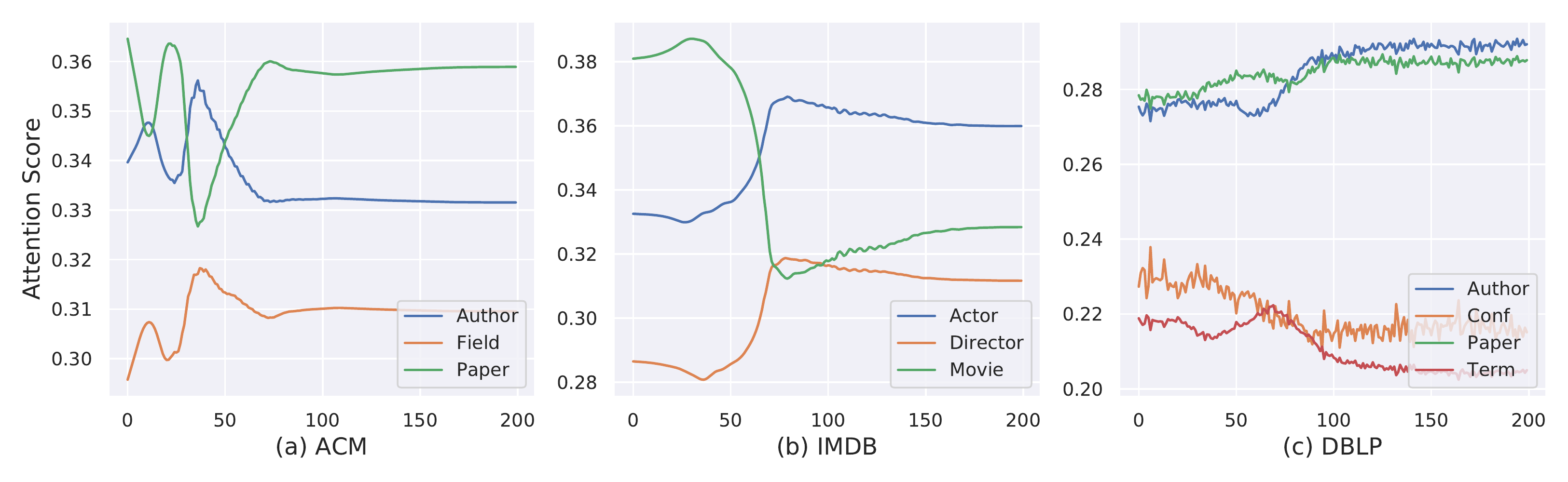}
	\caption{The attention values of node types w.r.t epochs on HGT.}
	\label{fig:att-hgt}
\end{figure*}
As mentioned earlier, we also analyze the attention value’s trend of HGT during its training process. Here, we conduct a detailed analysis of different datasets. AS can be seen in Figure~\ref{fig:att-hgt}(a), the attention values of Author, Field and Paper in ACM are slightly different with the increase of training epochs, but Paper basically maintains the maximum attention value. It proves that the node itself is significant for the paper classification task. From Figure~\ref{fig:att-hgt}(b), we notice that Movie has the largest attention value in IMDB, but as the training epoch increases, the attention value of Actor and Director becomes larger, and Movie is the opposite. It reveals that Movie is significant for movie classification tasks in the early stages of model training. However, as the model training time increases, Actors begin to play a leading role, followed by Movie. From Figure~\ref{fig:att-hgt}(c), the attention values of Author and Paper in DBLP gradually increase, while the attention values of Conf and Term are reversed. It means that Author and Paper are the most critical nodes that determine the author's research area. In summary, HIRE can fully reveal and weigh the differences between different nodes.

\begin{figure*}[htbp]
	\centering
	\includegraphics[width=\textwidth]{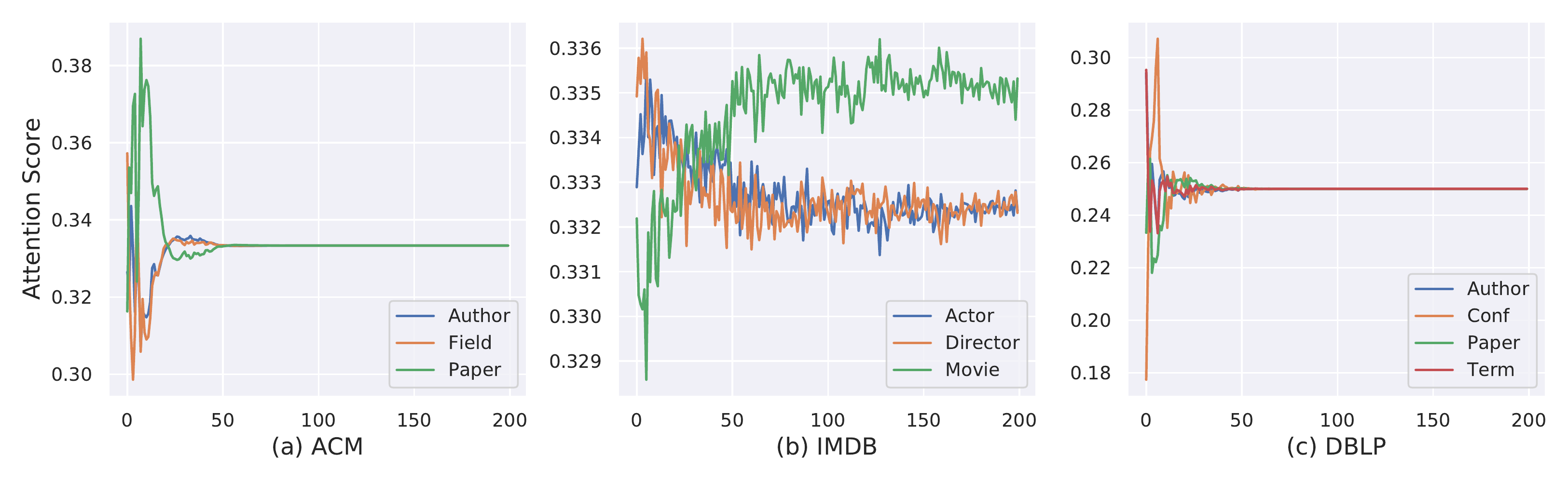}
	\caption{The attention values of node types w.r.t epochs on RGCN.}
	\label{fig:att-rgcn}
\end{figure*}
Similarly, we also analyze the changing trend of RGCN's attention value during the training process. As drawn in Figure~\ref{fig:att-rgcn}(a), Paper's attention value increases first and then decreases as the training epoch increases, but Field and Paper are the opposite. Finally, their attention values tend to be proximal, but Paper maintains the maximum attention value, demonstrating that the node is vital for the paper classification task. From Figure~\ref{fig:att-rgcn}(b), the attention value of Actor and Director becomes smaller, and the Movie becomes larger. It can be concluded that Actor and Director are the vital nodes for movie classification tasks in the early stages of model training. However, the Movie begins to play a leading role as the model training time increases. From Figure~\ref{fig:att-rgcn}(c), the attention values of Author, Paper, and Term decrease first and then increase, while Conf shows the contrary trend, and their values become proximal finally. It reflects that Conf determines the author's research area at first. As the model is further trained, Author, Paper, and Term used as supplementary information begin to assist the node classification task.

\clearpage


\bibliography{ref}
\newpage
    
\begin{minipage}[b]{0.35\linewidth}
    \includegraphics[height=8\baselineskip]{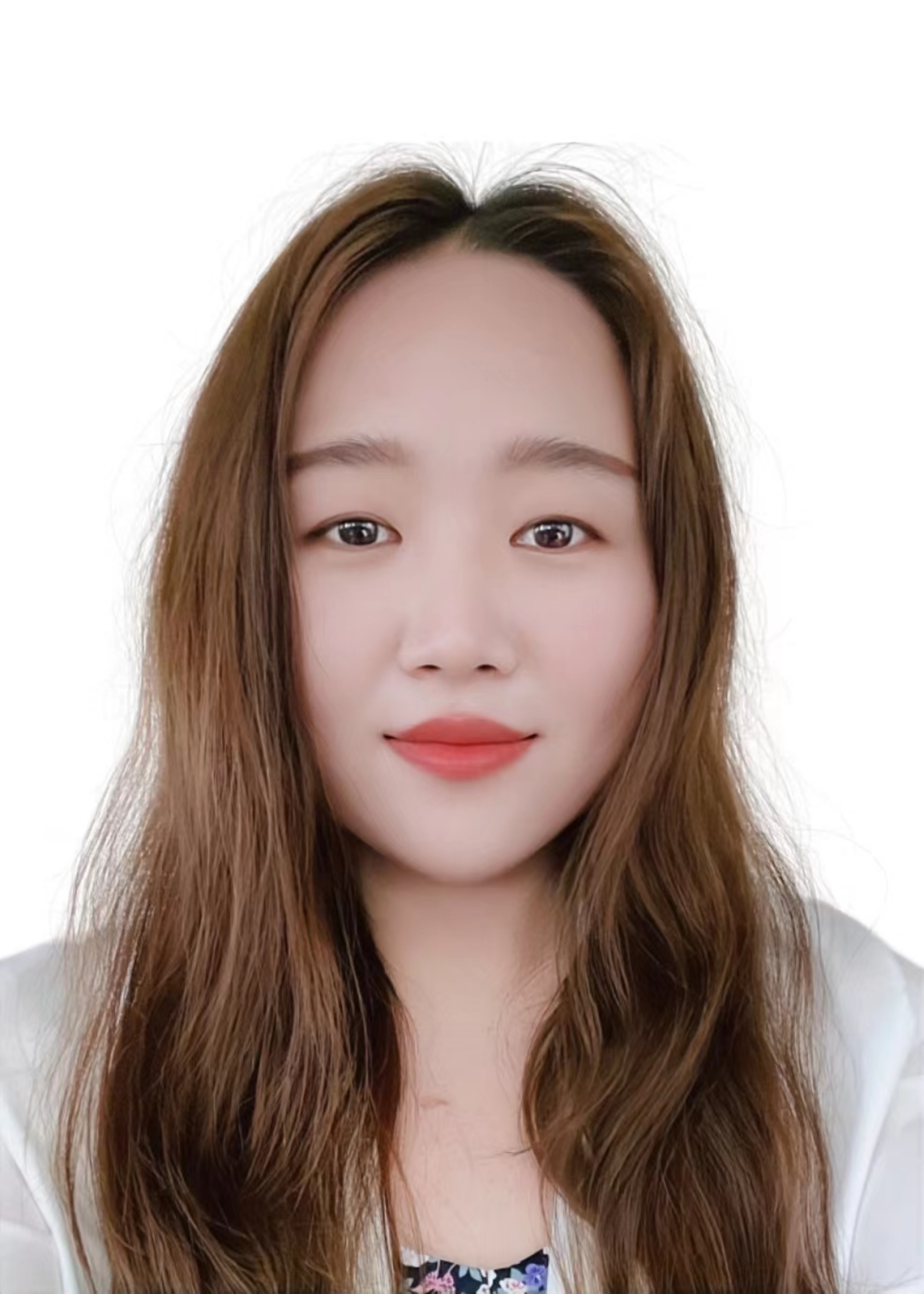}
\end{minipage}
\hfill
\begin{minipage}[b]{0.65\linewidth}
    \textbf{Jing Liu} is currently pursuing the Ph.D. degree with Institute of Computing Technology, Chinese Academy of Sciences. Her research interests include graph neural network, heterogeneous graphs representation learning and knowledge distillation.
\end{minipage}
\vskip 2cm

\begin{minipage}[b]{0.35\linewidth}
    \includegraphics[height=8\baselineskip]{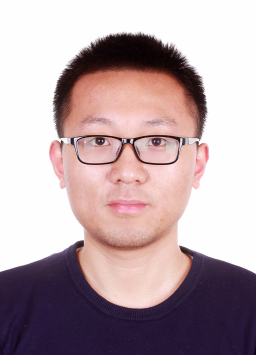}
\end{minipage}
\hfill
\begin{minipage}[b]{0.65\linewidth}
    \textbf{Tongya Zheng} is currently pursuing the Ph.D. degree with the College of Computer Science, Zhejiang University. His research interests include stream data computing and graph mining.
\end{minipage}
\vskip 2cm

\begin{minipage}[b]{0.35\linewidth}
    \includegraphics[height=8\baselineskip]{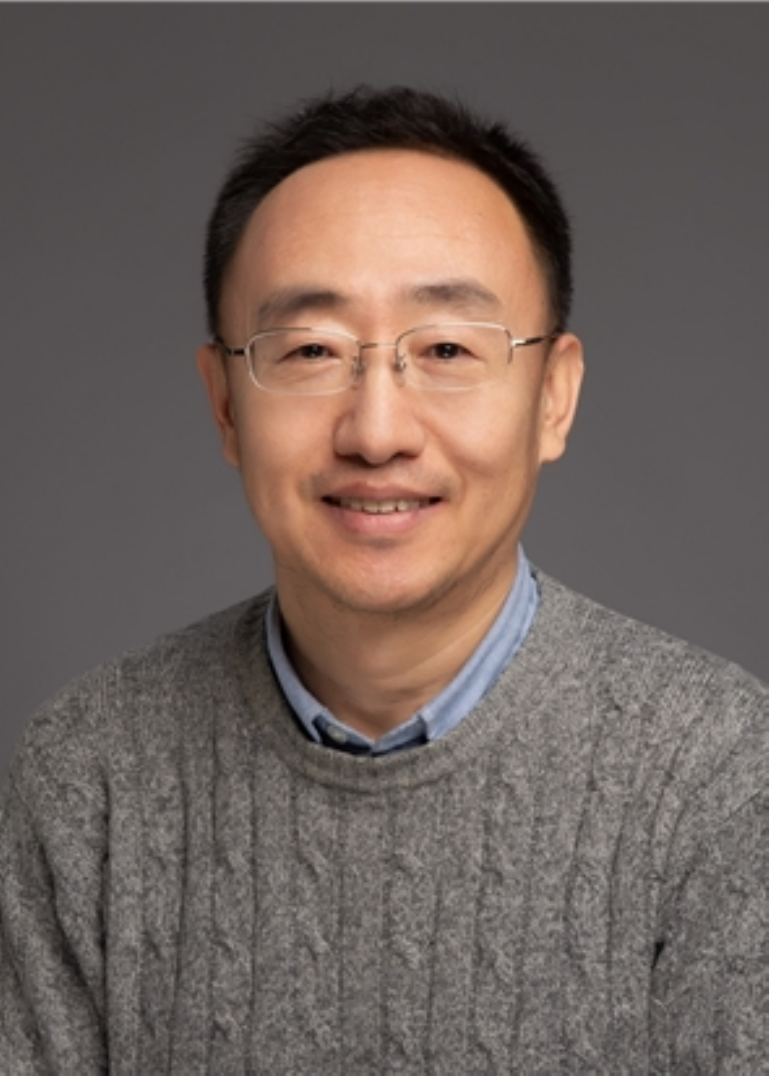}
\end{minipage}
\hfill
\begin{minipage}[b]{0.65\linewidth}
     \textbf{Qinfen Hao} received the Ph.D. degree in computer system architecture from the Institute of Computing Technology, Chinese Academy of Sciences, China, in 2006. He is currently a Researcher of Institute of Computing, Chinese Academy of Sciences. His research interests include computer architecture and graph computing.
\end{minipage}

\end{document}